%% file: acl_latex.tex
\pdfoutput=1

\documentclass[11pt]{article}

\usepackage[preprint]{acl}

\usepackage{times}
\usepackage{latexsym}

\usepackage[T1]{fontenc}

\usepackage[utf8]{inputenc}

\usepackage{microtype}

\usepackage{inconsolata}

\usepackage{graphicx}

\usepackage{hyperref}

\usepackage{colortbl}
\usepackage{booktabs}
\usepackage{multirow}

\usepackage{algorithm}
\usepackage{algpseudocode}
\usepackage{algorithmicx}
\usepackage{setspace}

\usepackage{caption}
\usepackage{subcaption}

\algnewcommand\algorithmicforeach{\textbf{for each}}
\algdef{S}[FOR]{ForEach}[1]{\algorithmicforeach\ #1\ \algorithmicdo}

\usepackage{amsmath}

\usepackage{cleveref}

\usepackage{xcolor}
\usepackage{tcolorbox}
\usepackage{listings}

\usepackage{times}
\usepackage{latexsym}
\usepackage{booktabs}
\usepackage{amsmath}
\usepackage{graphicx}
\usepackage{multirow}
\usepackage{tabularx}
\usepackage{enumitem}
\usepackage{svg}
\usepackage{tabularray}
\usepackage{longtable}
\usepackage{amsthm}
\usepackage{bm}
\usepackage{setspace}

\usepackage{pifont}
\newcommand{\cmark}{\ding{51}} 
\newcommand{\xmark}{\ding{55}} 

\newcommand{\promptcg}{\text{prompt}_{\text{GCG}}}
\newcommand{\llmcg}{\text{LLM}_{\text{GCG}}}
\newcommand{\llma}{\text{LLM}_{\text{A}}}
\newcommand{\codials}{$\text{CoDial}_{\text{structured}}$}
\newcommand{\codialf}{$\text{CoDial}_{\text{free}}$}
\definecolor{headercolor}{gray}{0.9}

\setitemize{labelindent=0em,labelsep=0.2cm,leftmargin=*}

\title{CoDial: Interpretable Task-Oriented Dialogue Systems\\ Through Dialogue Flow Alignment
}

\author{Radin Shayanfar\textsuperscript{1,2}, Chu Fei Luo\textsuperscript{1,2}, Rohan Bhambhoria\textsuperscript{1,2}, \\\textbf{Samuel Dahan\textsuperscript{2,3}, and Xiaodan Zhu\textsuperscript{1,2,4}}\\
\textsuperscript{1}Department of Electrical and Computer Engineering \& Ingenuity Labs, Queen's University\\
\textsuperscript{2}Conflict Analytics Lab, Queen's University\\
\textsuperscript{3}Cornell Law School \hspace{1.5mm}
\textsuperscript{4}Vector Institute for AI\\
\small{\{\texttt{radin.shayanfar,chufei.luo,r.bhambhoria,samuel.dahan,xiaodan.zhu}\}\text{\texttt{@queensu.ca}}}} 

\begin{document}
\maketitle
\begin{abstract}
Building Task-Oriented Dialogue (TOD) systems that generalize across different tasks remains a challenging problem. Data-driven approaches often struggle to transfer effectively to unseen tasks. While recent schema-based TOD frameworks improve generalization by decoupling task logic from language understanding, their reliance on neural or generative models often obscures how task schemas influence behaviour and hence impair interpretability.
In this work, we introduce a novel framework, \textbf{CoDial} (\underline{Co}de for \underline{Dial}ogue), at the core of which is converting a predefined task schema to a  structured heterogeneous graph and then to programmatic LLM guardrailing code, such as NVIDIA's Colang. The pipeline enables efficient and interpretable alignment of dialogue policies during inference. We introduce two paradigms for LLM guardrailing code generation, \codialf{} and \codials, and propose a mechanism that integrates human feedback to iteratively improve the generated code.
Empirically, CoDial achieves state-of-the-art (SOTA) performance on the widely used benchmark datasets, while providing inherent interpretability in the design. We additionally demonstrate CoDial's iterative improvement via manual and LLM-aided feedback, making it a practical tool for human-guided alignment of LLMs 
in unseen domains.\footnote{Our code and data are publicly available at \url{https://github.com/radinshayanfar/CoDial}.}

\end{abstract}

\input{sections/1_intro}
\input{sections/2_related}

\input{sections/3_method}

\input{sections/4_exp}
\input{sections/5_results}
\input{sections/6_conclusion}

\bibliography{custom}

\appendix

\input{sections/appendix}

\end{document}

%% file: sections/1_intro.tex
\section{Introduction}

\begin{figure}[t]

\setlength{\belowcaptionskip}{-3mm}
  \centering
  \includegraphics[width=\linewidth]{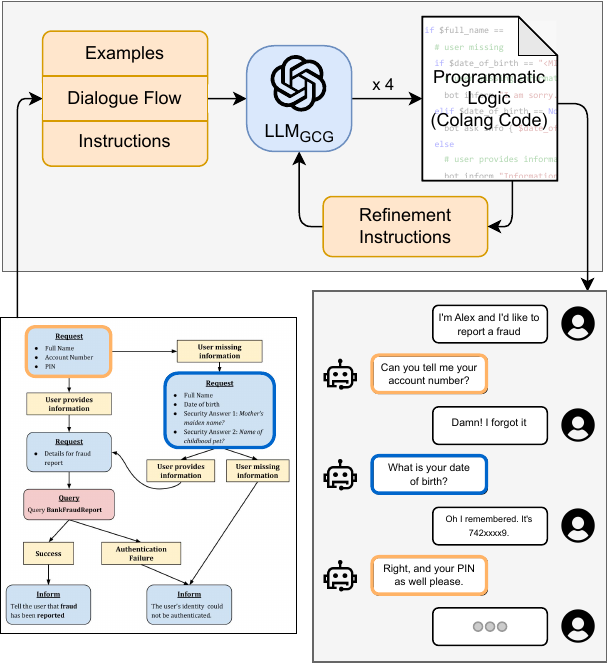}

   \caption{Overview of the proposed CoDial framework. An expert-curated dialogue flow (left) is transformed into executable programmatic logic using an LLM (top). The generated code is iteratively refined before producing the final program, which powers a conversational application (right), enabling the chatbot to follow the designer's requirements.}
   \label{fig:overview}
\end{figure}
Task-Oriented Dialogue (TOD) systems play a crucial role in a wide range of applications, enabling users to accomplish complex tasks such as flight booking or apartment searching through natural language conversation \citep{qin-etal-2023-end}. Building TOD systems that are capable of operating across different tasks remains a challenging area of exploration \citep{Doyoufollow-jacqmin-2022}. Data-driven approaches aim to train models on large corpora of conversations spanning multiple domains, allowing them to capture task-related conversational patterns. Such models often struggle with \textbf{generalization}: the ability to transfer effectively to new unseen task(s) \citep{SchemaGuidedParadigm-mehri-2021}.

Many recent TOD works adopt a schema-based approach to achieve zero-shot generalization, decoupling language understanding from task-specific dialogue policy \citep{SGPTODBuilding-zhang-2023, AnyTODProgrammableTask-zhao-2023, SchemaGuidedParadigm-mehri-2021}. These systems provide an approach to utilize a parsable \textit{task schema}, often represented as a graph, to encode and enforce complex task logic. 
Most schema-based methods rely on neural or fully generation-based parsing, which fall short on a key property: \textbf{interpretability}, or the capacity to examine how the schema is utilized by the model to produce specific outputs. In contrast to opaque neural or generative representations, a programmatic formulation allows one to inspect and reason about the decision process, thereby facilitating \textit{modification} and \textit{improvement} of the system.
Interpretability is also especially crucial in high-stakes domains such as law and medicine, where domain experts with minimal technical knowledge need to specify, validate, and refine AI behaviour \citep{dahan2023lawyers, tian2024opportunities}. 
Previous works \citep{AnyTODProgrammableTask-zhao-2023} design interpretability into the system by treating the task schema as a program to be executed by a language model. However, this approach requires humans to define the task programmatically, which typically demands greater effort and technical expertise than graph-based representations. This added requirement for programming expertise makes the approach less intuitive and increases the cost of adoption, particularly for non-technical users.

To enable \textit{generalizable, interpretable} TOD systems that adapt well to unseen tasks without requiring direct programming, we propose a novel framework, \textbf{CoDial} (\underline{Co}de for \underline{Dial}ogue). At the core of CoDial, we leverage programmatic Large Language Model (LLM) guardrailing languages, such as Colang \citep{nvidia2024nemo}. We reframe LLM guardrails as the foundation for defining TOD system behaviour. CoDial inherits the advantages of programmatic guardrailing, making the system interpretable by design and enabling flexible behaviour definition at inference time.

Specifically, we convert an input task schema, referred to as a \emph{dialogue flow}, into Colang code. We introduce two paradigms for generating programmatic guardrails: \codialf{} and \codials. Our key contributions include:

\begin{itemize}
\vspace{-1mm}
    \item We propose a novel approach for effective alignment of dialogue systems to unseen task schemas that is interpretable by design. To our knowledge, we are the first to treat TOD systems as programmatic LLM guardrailing, such as Colang code, and automate its generation. 

\vspace{-1mm}
    \item The proposed framework, CoDial, consists of three novel components. The heterogeneous dialogue flow representation provides a structure to define rich task schemas. The guardrail-grounded code generation pipeline transforms dialogue flows into executable LLM guardrailing programs, allowing for interpretable and flexible control of LLMs in the inference stage. The CoDial human-feedback mechanism incorporates human and LLM feedback to refine the generated guardrailed conversational models.

\vspace{-1mm}
    \item We demonstrate the effectiveness of our framework on publicly available TOD benchmarks, STAR and MultiWOZ. 
    The proposed pipeline achieves new state-of-the-art (SOTA) results on STAR and on par results with SOTA on MultiWOZ in a strict zero-shot setting.
    We also empirically evaluate the effect of different code refinement strategies, and provide a user study that illustrates CoDial's enhanced interpretability.

\end{itemize}

%% file: sections/2_related.tex
\section{Related Work}

\paragraph{Task-Oriented Dialogue}
While LLMs have demonstrated impressive capability in a wide variety of domains, they struggled with TOD and fell behind if not used properly \cite{AreLargeLanguage-hudecek-2023}. Some research \citep{SGPTODBuilding-zhang-2023, SchemaGuidedParadigm-mehri-2021} used a neural schema-guided approach to generalize TOD systems to unseen tasks without interpretability. AnyTOD \citep{AnyTODProgrammableTask-zhao-2023} provided an interpretable neuro-symbolic approach by viewing task schema as a manually-written policy program. However,

AnyTOD also relied on extensive training and exhibits limited generalization to unseen tasks.

\vspace{-0.5mm}
\paragraph{Guardrails} CoDial leverages guardrails to implement a TOD system. Guardrailing aims to enforce human-imposed constraints on LLMs at inference time~\citep{BuildingGuardrailsLarge-dong-2024,NeMoGuardrailsToolkit-rebedea-2023,guardrailsai}. While originating from AI safety, we argue that they can generally be used to define any desired behaviour of LLMs.
NVIDIA NeMo-Guardrails \citep{NeMoGuardrailsToolkit-rebedea-2023} is a toolkit that adds programmable guardrails to LLM-based conversational applications and employs Colang \cite{nvidia2024nemo}, a programming language, to establish highly flexible conversational flows.

\vspace{-0.5mm}

\paragraph{Code Generation and Prompt Optimization}

Code generation has made remarkable progress with the introduction of LLMs \citep{NEURIPS2022_8636419d}. Although there are still challenges, such as logical consistency and hallucinations \citep{liu2024your}, LLMs are proficient when in-context examples, documentation, or plans are provided \citep{jiang2024self}. There has been research to improve output by rewriting the input prompt, referred to as prompt optimization \citep{yuksekgonul2024textgrad}. Please refer to \Cref{ap:rw} for detailed related work.

%% file: sections/3_method.tex
\section{Methodology}

We introduce CoDial, a novel framework for constructing interpretable TOD systems without requiring training data or manual programming, as illustrated in \Cref{fig:overview}. A task schema, defining the behaviour of the TOD system, is the \emph{only} input of CoDial. The core of our approach is leveraging programmatic LLM guardrailing, which allows interpretable and flexible control over the behaviour of an LLM in the inference stage.

CoDial is composed of three key components: (1) CoDial Heterogeneous Dialogue Flows (\textbf{CHIEF}) that provides a framework to represent the predefined task schema (\Cref{sec:df-framework}), (2) Guardrail-Grounded Code Generation (\textbf{GCG}) that automatically creates a TOD system driven by an executable guardrailing program based on the input dialogue flow (\Cref{sec:code-gen}), and (3) CoDial Human Feedback (\textbf{CHF}) that incorporates human/LLM feedback to optimize the generated guardrailing application (\Cref{sec:code-optimization}). In this paper, we investigate two code generation paradigms for GCG and use the Colang~\cite{nvidia2024nemo} guardrailing language, but any other programmatic paradigm can be applied.

\subsection{CoDial Dialogue Flow Representation} \label{sec:df-framework}

We design a structured framework to represent rich task schemas, referred to as ``\emph{dialogue flows}'', as heterogeneous directed graphs, called \underline{C}oDial \underline{H}eterogeneous d\underline{I}alogu\underline{E} \underline{F}lows
(\textbf{CHIEF}) representation. 
Unlike prior work \citep{SchemaGuidedParadigm-mehri-2021,SGPTODBuilding-zhang-2023} that define the task schema as a homogeneous graph---where the single node type represents user intent, an API return value, or a dialogue state---CHIEF allows for different node or edge types in a heterogeneous manner, supporting structured and richer task definition (e.g., \Cref{fig:example-df}). 
To the best of our knowledge, we are the first to frame TOD task schema as a heterogeneous directed graph and structure its definition. Specifically, CHIEF provides different node types that can define rich metadata and natural language logic to cover a wide range of tasks and domains, inspired by \citet{STARSchemaGuided-mosig-2020}. \footnote{In this work, we used GPT-4o to convert an input homogeneous task schema into our CHIEF representation. Future unseen tasks can follow a similar method, or work directly with our CHIEF framework to rigorously define the logic.}
Below, we briefly discuss the main node types and actions in CHIEF. Refer to \Cref{ap:det-chief} for more details.
\paragraph{Request}  
The request nodes define variables, hereby referred to as slots, that CoDial tracks throughout the conversation (e.g. \textit{the departure location in a taxi booking task}). When a conversation reaches this node, the system will request information specified by the slots. Each slot is accompanied by a few example values and includes a free-form \texttt{rule} property to define the conditions under which a slot should be requested.
\paragraph{External Action}
This node specifies a call to an external function within a dialogue flow. This enables the designer to execute complex logics through programming functions, API interactions, or invoking an LLM.

\paragraph{Inform (and Confirm)}
This node defines a template for providing information to the user (e.g. \textit{Your taxi is booked with reference number [ref\_no]}), and an optional follow-up question (e.g. \textit{Do you confirm the booking?}).

\paragraph{Global and Fallback Actions}
CHIEF supports global and fallback actions that are not tied to particular dialogue steps. Global actions can be triggered at any point in the dialogue flow (e.g. responding to a greeting). We also define fallback actions, general responses used when no other action is selected (e.g. \textit{Sorry, I can't help with that}).

The defined nodes logically connect with \textbf{edges}. We add a textual \texttt{condition} property to edges to allow conditional branching in dialogue flows. We encode the graphs defined by CHIEF as text in JSON format (e.g., \Cref{fig:example-json}). The JSON-encoded representation is translated into programmatic guardrails with GCG, described below.

\subsection{Guardrail-Grounded Code Generation}

Guardrailing is a general paradigm to define the flow of conversational systems and enable inference-stage control over LLMs' behaviour \citep{BuildingGuardrailsLarge-dong-2024,NeMoGuardrailsToolkit-rebedea-2023}. Unlike neural models, programming codes are inherently interpretable. Therefore, programmatic guardrailing allows interpretable and flexible behaviour definition in conversational systems.
Our work is the first to formulate TOD system as programmatic guardrailing and automate its generation, removing the need and technical barrier of programming while ensuring interpretability.

We propose CoDial Guardrail-Grounded Code Generation (\textbf{GCG}) that translates CHIEF representations into guardrailing code (e.g. Colang\footnote{\url{https://docs.nvidia.com/nemo/guardrails/latest/configure-rails/colang/colang-2/index.html}}). GCG is performed by prompting a code generation model, $\llmcg$, with detailed specifications $\promptcg$\footnote{We also experimented with (1) retrieval-augmented generation using the Colang Language Reference documentation and (2) fine-tuning GPT-4o-mini on generation pairs of $(\text{programming task}, \text{Colang code})$, but found that prompting with examples works best.}.
Formally, the GCG process is denoted as $g=\llmcg\left(\promptcg\left(x\right)\right)$, where $\promptcg\left(x\right)$ is a JSON-encoded CHIEF graph $x$ wrapped with the prompt template instructions, and $g$ is the program that guardrails the dialogue LLM agent, $\llma$. 

We investigate two different paradigms for implementing $\promptcg$ in GCG. In the first paradigm, denoted as \codialf, $\promptcg$ provides LLM with the syntax and semantic rules of the guardrailing language.
Because several code implementations may be valid for a given problem, this paradigm leaves $\llmcg$ free to design a guardrailing logic that models the given dialogue flow.
The second paradigm directly instructs LLM with a certain dialogue flow modelling approach, specifying the structure of $g$ and how to manage the dialogue, interpret each CHIEF node, and implement its equivalent guardrailing code. We denote the latter approach as \codials. \Cref{fig:example-code-2,fig:example-code,fig:example-comparison} illustrate a dialogue flow, its JSON schema representation in CHIEF, and the Colang code generated by \codialf{} and \codials{}. Please refer to \Cref{ap:det-cg} for more details on code generation.

\subsubsection{\codialf}

Since most LLMs are unfamiliar with guardrailing languages, we include the documentation of our chosen language, Colang, in $\promptcg$. As a preliminary design and due to the large context of the documentation, we hand-pick the most essential chunks to provide $\llmcg$ with a general understanding of Colang's syntax and semantics.

\Cref{fig:prompt-overview2} illustrates an overview of the $\promptcg$ for \codialf. The prompt begins with Colang syntax and semantic rules, followed by the input dialogue flow $x$, and concludes with a task description instructing the model to generate Colang code for the flow. The generated code $g$ is an executable guardrailing program that specifies a TOD system aligned to the given CHIEF representation. We also instruct $\llmcg$ to enable Colang's \texttt{continuation on unhandled user intent} flow to allow $\llma$ to generate output, given fallback actions and all actions defined in the dialogue flow, if the guardrails do not match with the user input in a conversation turn. \Cref{fig:example-code-2} shows an example of a generated code in \codialf.

 \begin{algorithm}[t]
\algrenewcommand\alglinenumber[1]{\scriptsize #1:}
\footnotesize
\caption{An outline of \codials.}\label{alg:main}

\begin{spacing}{1.1}
\begin{algorithmic}[1]

    \ForEach{$v^{(H)}$ in $V^{(H)}$}
        \State $v^{(H)} \gets$ \textsc{null} or \textsc{false} \Comment{Init helper variables}
    \EndFor
    \While{True}
        \State $h_{2i-1} \gets \left(h_{2i-2};\, U_i\right)$ \Comment{Append user input to history}

        \Statex
        \State \texttt{intent} $\gets$ \Call{DetectIntent}{$U_i$} \Comment{Global action}
        \If{\texttt{intent} $\neq$ \textsc{null}}
            \State $B_i \gets$ \Call{IntentResponse}{\texttt{intent}}
            \State \textbf{continue}
        \EndIf

        \Statex
        \ForEach{$v^{(S)}_j$ in $V^{(S)}$} \Comment{DST -- update all slots}
            \State $v^{(S)}_{old} \gets v^{(S)}_j$
            \State $v^{(S)}_j \gets \text{DST}\left(h_{2i-1},\, p^{(S)}_j,\, \llma\right)$
            \If{$v^{(S)}_j \neq v^{(S)}_{old}$}
                \State $V^{(H)}_j \gets \{ v \in V^{(H)}\mid \exists\, e = (v^{(S)}_j, v)\}$ 
        \vspace{0.1mm}        \Statex\Comment{Find dependent helper variables}
        
        \vspace{0.1mm}                \ForEach{$v^{(H)}$ in $V^{(H)}_j$}
                    \State $v^{(H)} \gets$ \textsc{null} or \textsc{false}
                \EndFor
            \EndIf
        \EndFor
        \Statex
        \State \texttt{state} $\gets$ $\left(V^{(S)},\, V^{(H)}\right)$
        \State $a_i \gets$ \Call{NAP}{\texttt{state}, $\llma$} \Comment{NAP}
        \State $V^{(H)}_{\texttt{state}} \gets \{ v^{(H)} \in V^{(H)} \mid \text{node}(v^{(H)}) = \text{node}(a_i) \}$ 
        \Statex\Comment{Update helpers at predicted node}
        \vspace{2mm}        \ForEach{$v^{(H)}$ in $V^{(H)}_{\texttt{state}}$}
            \State $v^{(H)} \gets$ \textsc{true} \textbf{if} $v^{(H)} =$ \textsc{false} \textbf{else} \Call{ExternalAction}{$v^{(H)},\, \texttt{state}$}
        \EndFor
        \Statex
        \If{$a_i =$ \textsc{null}} \Comment{Fallback action}
            \State $B_i \gets \llma\!\left(V^{(S)},\, V^{(H)}\right)$
        \Else
            \State $B_i \gets a_i$
        \EndIf
    \EndWhile
\end{algorithmic}
\end{spacing}

\end{algorithm}

\subsubsection{\codials}
\label{sec:code-gen}

The simple design of \codialf{} serves as an interpretable baseline where LLMs generate TOD programs from CHIEF representations and language documentation without guidance.
Additionally, we propose \codials, where we explicitly instruct the model on how to structure the code, model the dialogue states, and interpret each CHIEF node type for GCG.
\Cref{fig:prompt-overview} shows an overview of $\promptcg$ for \codials.

\begin{figure}[t]
  \centering
  \includegraphics[width=.7\linewidth]{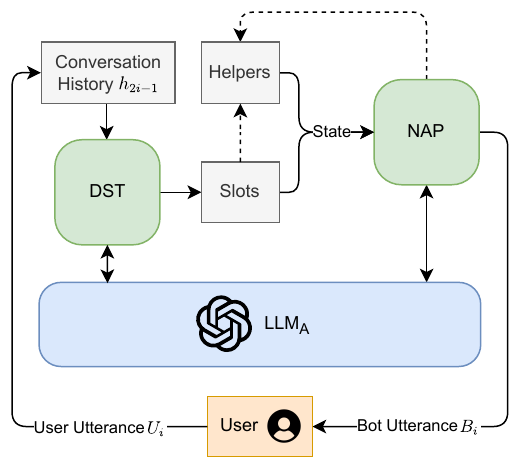}

   \caption{Execution life cycle of the generated agent in \codials.}
   \label{fig:runtime}
\end{figure}

Our $\promptcg$ outlines the output guardrailing code $g$, as presented in \Cref{alg:main}. We define notations later in this section. The conversation runs within an infinite \texttt{while} loop, where the TOD system (1) waits for user input, (2) detects the user’s intent for global actions, (3) predicts the slot variables (Dialogue State Tracking; DST); (4) selects an action and generates a response (Next Action Prediction; NAP). In this work, we leverage Colang's built-in intent detection feature for global actions. Note that DST and NAP are combined in a single executable program (i.e., $g$).
Finally, if the NAP component does not generate a response to the given user utterance (e.g., the conversation strays from the defined logic), $\llma$ is directly prompted to choose from all available actions, including fallbacks, based on the conversation history. 
\Cref{fig:runtime} shows the full execution life cycle.

We denote a conversation between user $U$ and chatbot $B$ as a history of messages, Equation \ref{eq:history}, where $U_{i}$ and $B_{i}$ show user's and chatbot's $i$-th utterance, respectively. Therefore, the total number of utterances in a conversation history $h_{2i}$ is $2i$.
\begin{equation} \label{eq:history}
    h_{2i} = \\
    \left( U_1 , B_1, \dots, U_{i}, B_{i} \right)
\end{equation}

We define a set of slot variables $V^{(S)}$ that track values for all of the slots defined in request nodes, and helper variables $V^{(H)}$ that track the state for other (non-request) types of nodes. The union of $V^{(S)}$ and $V^{(H)}$ forms the state of conversation $s = \left( V^{(S)}; V^{(H)} \right)$ at each turn, which is used to determine the next action.

\paragraph{Dialogue State Tracking (DST)} \label{sec:dst}
As suggested by \citet{TowardsLLMdriven-feng-2023}, 
LLM prompting shows promising performance in DST, so we take a similar prompting approach in this work.
For each slot, $\llmcg$ creates explicit instructions to extract the value from the entire conversation history. We leverage Colang's Natural Language Description (NLD) feature to execute these instructions with $\llma$ and save the value to a slot variable.
Formally, a slot variable $v^{(s)}_j \in V^{(S)}$ is predicted as Equation \ref{eq:dst}, where $p^{(s)}_j \in P^{(s)}$ is the prompt generated by $\llmcg$ to extract the value for $v^{(s)}_j$.

\begin{equation} \label{eq:dst}
    v^{(s)}_j = \text{DST}\left(h_{2i-1}, \llma, p^{(s)}_j \right)
\end{equation}

\paragraph{Next Action Prediction (NAP)} \label{sec:nap}
We instruct $\llmcg$ to convert the CHIEF graph $x$ into a conditional logic consisting of nested if/else statements, to generate a response given $s_i$, the state of the conversation at turn $i$. Each generated if statement corresponds to a node $n_j$ in $x$, and aligns $s_i$ to the conversation logic outlined by the CHIEF representation. If an if statement holds, this indicates $s_i$ is ``at'' that node and the corresponding action is executed; otherwise, for each outgoing edge at node $n_j$, the system checks for traversal. If there is a natural language condition associated with the edge and the condition is met, or if there is no explicit condition, $\llma$ traverses the graph to the associated target node. Formally, next bot utterance is defined in Equation \ref{eq:nap}.
\begin{equation} \label{eq:nap}
    \left(B_{i}, V^{(H)}_{i+1} \right) = \text{NAP}(s_i, \llma)
\end{equation}

\subsection{CoDial Human Feedback Integration} \label{sec:code-optimization}

CoDial's Human Feedback (CHF) mechanism incorporates human feedback to refine the generated guardrailing code $g$. The code enhancement through feedback comprises two broad approaches: i) manual and ii) LLM-aided modifications.

CHF assists iterative improvement of $g$ in the form of \textbf{refinement instructions (RIs)}, shown at the top in \Cref{fig:overview}. RIs allow the user of CoDial to refine the generated logic through natural language. We provide three instructions for refining the output code: correct logic (i.e., if statement) for each node, DST initialization, and request node checks. 
Since these RIs, presented in \Cref{tab:ref-ins}, are a set of prompts, they can be modified and extended dynamically.
In addition, CHF allows for manual modifications on the dialogue flow (\Cref{ap:star}) and manual DST prompt optimization (\Cref{ap:mwz}).
We also experiment with automatic prompt optimization, detailed in \Cref{ap:mwz}.

%% file: sections/4_exp.tex
\section{Experimental Settings}
\vspace{-1mm}

\paragraph{Models}
We use GPT-4o, GPT-5 (with reasoning levels of \underline{m}inimal and \underline{l}ow), Claude 3.5 Sonnet, Gemini 2.0 Flash, Qwen3-30B-A3B, and DeepSeek V3 (DSV3) as $\llmcg$ and $\llma$. Larger models are used for code generation---given the complexity of the task, we found that smaller models often fail to fully adhere to instructions. For further details, please refer to \Cref{ap:exp}.

\vspace{-0.5mm}
\begin{table*}[t]
    \centering
    \setlength{\tabcolsep}{8pt}
    \renewcommand\arraystretch{1.05}
    \resizebox{0.95\linewidth}{!}{
    \begin{tabular}{l c c | 	*{3}{>{\centering\arraybackslash}p{1.3cm}} | *{5}{>{\centering\arraybackslash}p{1.3cm}} }
        \toprule
    \multirow{2}{*}[-1mm]{\textbf{Model}} & \multirow{2}{*}[-1mm]{\textbf{Int.}} & \multirow{2}{*}[-1mm]{\textbf{\begin{tabular}[c]{@{}c@{}}Graph\\ Transfer\end{tabular}}}  & \multicolumn{3}{c}{\textbf{STAR}} & \multicolumn{5}{c}{\textbf{MultiWOZ 2.2}} \\
        \cmidrule(lr){4-6} \cmidrule(lr){7-11}
        & & & F1 & Accuracy & BLEU & JGA & Inform & Success & BLEU & Combined \\
        \midrule
SOLOIST & \xmark & \xmark & - & - & - & 35.9         & 81.7            & 67.1             & 13.6          & 88.0                \\
MARS & \xmark & \xmark & - & - & - & 35.5         & 88.9            & 78.0               & \underline{19.6}          & 103.0            \\
DARD & \xmark & \xmark & - & - & - & - & \underline{96.6}         & \underline{88.3}            & 12.1             & \underline{104.6} \\
IG-TOD {\small{}(few-shot)} & \xmark & \xmark & - & - & - & 27           & -               & 44               & \underline{6.8}          & -                 \\
        \cmidrule(lr){4-11}
        \multicolumn{3}{l|}{} & \multicolumn{8}{c}{\cellcolor{headercolor}(Strict) Zero-shot Transfer} \\
IG-TOD {\small{}(zero-shot)} & \xmark & \xmark & - & - & - & 13           & -               & 31               & 4.2          & -                 \\
BERT + Schema & \xmark & \cmark & 29.7* & 32.4* & - & - & - & - & - & - \\
SAM & \xmark & \cmark & 51.2* & 49.8* & - & - & - & - & - & - \\
AnyTOD \textsc{xxl} & \cmark & \xmark & \underline{68.0}* & \underline{68.0}* & 44.3* & 30.8         & 76.9            & 47.6             & 3.4           & 65.6 \\ 
        SGP-TOD & \xmark & \cmark & 53.5 & 53.2 & -            & - & \textbf{82.0}              & \textbf{72.5}             & \textbf{9.2}          & \textbf{86.5} \\
        \midrule
        \textbf{\codialf} & \cmark & \cmark & \multicolumn{8}{l}{} \\
~~CoDial {\small{}(4o, 4o-mini)} $-$ \textsc{ri} &  &  & 36.6 & 36.1 & 23.0 & - & - & - & - & - \\ 
\midrule
        \textbf{\codials} & \cmark & \cmark & \multicolumn{8}{l}{} \\
~~CoDial {\small{}(4o, 4o-mini)}       &  &  & 58.5 & 60.1 & 45.2 & 28.4        & 76.6              & 54.6            & 3.5         & 69.1 \\
~~CoDial {\small{}(4o, 5-mini:l)}       & & & \textbf{59.2} & \textbf{60.2} & \textbf{46.5} & \textbf{37.0}        & \underline{79.6}              & \underline{70.8}            & 4.3         & \underline{79.5} \\
        \bottomrule
    \end{tabular}}
    \caption{Comparison of CoDial with baselines on STAR and MultiWOZ benchmarks.
    In ``Strict Zero-Shot'' the models have not seen a same task schema architecture in the training data. Results with an asterisk (*) are evaluated in a more relaxed, non-strict setting, and therefore, are not directly comparable. ``Int.'' stands for ``Interpretable.'' SAM results are cited from \citet{AnyTODProgrammableTask-zhao-2023}.} 
    \label{tab:star-mwz}
\end{table*}

\subsection{Datasets}

\vspace{-0.5mm}

\paragraph{STAR}
The STAR dataset \citep{STARSchemaGuided-mosig-2020}, collected in a Wizard-of-Oz setup (2,755 human-human conversations), provides \textbf{explicit task schemas} (i.e., dialogue flows) to ensure consistent and deterministic system actions. 
It serves as a benchmark for TOD systems, enabling evaluation across 24 tasks and 13 domains. STAR's structured collection aligns well with our objectives and CoDial's design choices. We also use silver state annotations created in STARv2 \citep{AnyTODProgrammableTask-zhao-2023} for ablations.
Refer to \Cref{ap:star} for more implementation details.

\vspace{-0.5mm}
\paragraph{MultiWOZ}

MultiWOZ \citep{MultiWOZLargeScale-budzianowski-2018} is a large-scale, multi-domain TOD dataset consisting of 1,000 human-human conversations, with most domains involving booking subtasks such as hotel reservations and taxi services.
Since MultiWOZ does not provide explicit dialogue flows, we manually construct them by analyzing example dialogues from each domain.
Given the impracticality of crafting dialogue flows for every possible domain combination \citep{SGPTODBuilding-zhang-2023}, we report results in a naive oracle domain setting.
Please refer to \Cref{ap:mwz} for more details.

\subsection{Metrics}
\vspace{-0.5mm}

For the STAR dataset, we compute BLEU-4 score \citep{BLEUmethodautomatic-papineni-2002, CallClarityReporting-post-2018} and follow \citet{STARSchemaGuided-mosig-2020} to compute F1 and accuracy. 
For the MultiWOZ dataset, we compute BLEU, Inform and Success rates, and Joint Goal Accuracy (JGA) using the official evaluation script \citep{ShadesBLEUFlavours-nekvinda-2021}. Since \codialf{} does not include an explicit DST component and most MultiWOZ metrics rely on DST predictions, we do not report \codialf{} results on this dataset.
We report the mean of three runs.

\subsection{Baselines}

For a complete list of compared methods, please refer to \Cref{ap:baselines}.
Our most comparable baselines are as follows:
\begin{itemize}
\vspace{-1mm}
    \item \textit{IG-TOD} \citep{AreLargeLanguage-hudecek-2023} is a prompting-based approach using ChatGPT to track dialogue states via slot descriptions, retrieve database entries, and generate responses without fine-tuning.
\vspace{-2mm}
    \item \textit{AnyTOD} \citep{AnyTODProgrammableTask-zhao-2023} pretrains and fine-tunes T5-XXL for dialogue state tracking and response generation. It uses a Python program to enforce the complicated logic defined by a dialogue flow to guide the LM decisions.
\vspace{-2mm}
    \item \textit{SGP-TOD} \citep{SGPTODBuilding-zhang-2023} is a purely generative approach that uses two-stage prompting to track dialogue state and generate response. It employs graph-based dialogue flows to steer LLM actions without requiring fine-tuning or training data. Refer to \Cref{ap:baselines} for details on fair comparison.
\vspace{-2mm}
    \item \textit{BERT + Schema} and \textit{Schema Attention Model (SAM)} \citep{STARSchemaGuided-mosig-2020,SchemaGuidedParadigm-mehri-2021} incorporate task schemas by conditioning on the predefined schema graphs, enabling structured decision-making in TODs. Both models rely on fine-tuning to learn schema-based task policies and improve generalization across tasks.
\end{itemize}
\vspace{-1mm}
For the remainder of this paper, by CoDial we refer to \codials{} with GPT-4o and GPT-4o-mini as $\llmcg$ and $\llma$, respectively, unless otherwise specified.

%% file: sections/5_results.tex
\section{Experimental Results}
\label{sec:results}

\label{sec:star}

\paragraph{Superior Performance with Explicit Schemas} 
\Cref{tab:star-mwz} summarizes CoDial results on the benchmark datasets.
CoDial achieves strong performance, surpassing all baselines except AnyTOD, and sets the new SOTA in strict zero-shot setting, where no same-architecture task schema is seen by the model. Our framework improves F1 by $\textbf{+5.7}$ and accuracy by $\textbf{+7}$ points over the previous SOTA. While AnyTOD achieves higher scores, it is evaluated in the easier non-strict setting and requires the task designer to write code, limiting accessibility to non-programmers. In contrast, CoDial operates in a graph-based transfer manner, eliminating the need for manual programming. 
We also observe that \codialf{} lags behind \codials{} and most baselines, indicating that LLMs struggle with unsupervised guardrailing code generation, likely due to limited availability of guardrailing languages, and still require human supervision.

\paragraph{Competitive Performance on MultiWOZ}
\label{sec:star-mwz}

\Cref{tab:star-mwz} also shows our results on the MultiWOZ dataset.
Unlike STAR, where wizards were provided structured guidance for system responses, MultiWOZ lacks a predefined dialogue flow, making interactions less consistent. 
This variability in MultiWOZ poses additional challenges for heuristics-grounded and programmatic approaches like CoDial and AnyTOD. Consequently, CoDial is less effective on MultiWOZ. 
To address this, we experiment with GPT-5 with built-in reasoning as $\llma$ to improve the DST performance. We observe that CoDial achieves competitive performance with SOTA on Inform and Success metrics under the strict zero-shot setting, while maintaining interpretability. We further analyze the effect of DST performance in \Cref{sec:ablations}.
Similar to AnyTOD, CoDial relies on template-based outputs, which accounts for its lower BLEU score.

\subsection{Detailed Analysis}

\paragraph{Impact of Model Selection and CHF}

We experiment with different model choices for the ($\llmcg$, $\llma$) pairing (\Cref{tab:star-only}). 
Better instruction following and more robust code generation often translate to higher overall performance. Because most LLMs are unfamiliar with guardrailing languages such as Colang, they must accurately interpret the $\promptcg$ to produce syntactically correct code. 
When the chosen LLM struggles with instruction following, code generation can fail, leading to incorrect or incomplete programs. 
Among the tested configurations, CoDial \textsc{(4o, 5-mini)} with built-in reasoning achieve the highest performance in all metrics. CoDial \textsc{(4o, 4o-mini)} performs comparably with lower cost and latency. Therefore, we use GPT-4o-mini for our ablations in \Cref{sec:ablations}.
We also report results in an oracle voting setting (\Cref{tab:star-abl}) between GPT-4o-mini and DSV3 as $\llma$, where for each task, we take the best-performing $\llma$ by F1. This results in an increase of $+1.7$ F1 and $+1.5$ accuracy.

Additionally, without modifications (\Cref{ap:star}), the original STAR dialogue flows result in lower performance (F1: $51.9$). 
Manually modifying the CHIEF representation and applying RIs to the generated code significantly enhances performance. We further explore the impact of LLM-aided corrections in \Cref{sec:ablations}.

\begin{table}[t]
    \centering
    \setlength{\tabcolsep}{8pt}
    \renewcommand\arraystretch{1.05}
    \resizebox{0.99\linewidth}{!}{
    \begin{tabular}{l | 	*{2}{>{\centering\arraybackslash}p{1.55cm}} *{1}{>{\centering\arraybackslash}p{0.5cm}} | *{3}{>{\centering\arraybackslash}p{0.8cm}} }
        \toprule
\textbf{Model}  & \textbf{$\llmcg$} & \textbf{$\llma$} & \textbf{RI} & \textbf{F1}      & \textbf{Acc.}    & \textbf{BLEU} \\        \midrule
        \rowcolor{headercolor}        \multicolumn{7}{l}{\textbf{\codialf}} \\
~~CoDial $-$ \textsc{ri}  & 4o                & 4o-mini  & \xmark         & 36.6             & 36.1               & 23.0          \\ 
~~CoDial $-$ \textsc{ri}  & Sonnet                & 4o-mini  & \xmark         & 32.1             & 31.8               & 18.0          \\ \midrule
        \rowcolor{headercolor}        \multicolumn{7}{l}{\textbf{\codials}} \\
~~CoDial \textsc{\small{}original DFs} & 4o                & 4o-mini  & \cmark         & 51.9             & 51.1               & 38.9          \\
~~CoDial $-$ \textsc{ri}          & 4o                & 4o-mini & \xmark        & 56.1             & 57.3             & 38.4          \\
~~CoDial $-$ \textsc{ri}          & Sonnet            & 4o-mini & \xmark        & 57.0                 & 58.4                 & 39.2              \\~~CoDial   & DSV3              & 4o-mini & \cmark        & 46.1                 & 48.0                 & 28.0              \\
~~CoDial*               & Gem. {\small{}2 Fl.}         & 4o-mini & \cmark        & 50.5                 & 52.1          & 32.9              \\
~~CoDial                  & Sonnet            & 4o-mini & \cmark        & 57.7                 & 58.5                 & 39.3              \\
~~CoDial                  & 4o                & Qwen 3 & \cmark           & 55.1           & 57.4             & 23.04              \\
~~CoDial                  & 4o                & DSV3 & \cmark           & 55.6           & 56.8             & 44.2              \\
~~CoDial         & 4o                & 4o-mini & \cmark        & 58.5    & 60.1    & 45.2 \\
~~CoDial         & 4o                & 5-mini:m & \cmark        & 59.0    & \textbf{60.4}    & 45.2 \\
~~CoDial         & 4o                & 5-mini:l & \cmark        & \textbf{59.2}    & 60.2    & \textbf{46.5} \\
        \bottomrule
    \end{tabular}}
    \caption{Comparison of CoDial performance across different settings and ($\llmcg$, $\llma$) pairs on STAR dataset. The generated code for the model with an asterisk (*) has been manually fixed and is not directly comparable. DF stands for ``dialogue flow.''} 
    \label{tab:star-only}
\end{table}

\begin{table*}[t]
\captionsetup{skip=7pt}
\centering
\begin{minipage}[t]{0.31\textwidth}
\strut\vspace*{-\baselineskip}\newline    \centering
    \setlength{\tabcolsep}{12pt}
    \resizebox{.8\columnwidth}{!}{
    \renewcommand\arraystretch{1.05}
    \begin{tabular}{l|c|c}
    \toprule
    \textbf{Actions} & \textbf{F1} & \textbf{Acc.} \\
    \midrule
    \rowcolor{headercolor}    \multicolumn{3}{l}{\textit{Intent Detection (Global Actions)}} \\
        All & 96.3 & 92.8 \\
    \midrule
    \rowcolor{headercolor}    \multicolumn{3}{l}{\textit{LLM Generated Actions}} \\
        Fallbacks & 51.4 & 57.8 \\
        Excluding Fallbacks & 38.7 & 39.0 \\ 
        All & 49.1 & 52.1 \\ \bottomrule
    \end{tabular}}
    \caption{Individual action prediction performance of intent detection and $\llma$ in CoDial. Fallback actions include \texttt{goodbye}, \texttt{out\_of\_scope}, and \texttt{anything\_else}. All entries are micro-averaged.}
    \label{tab:llm-gen-perf}
\end{minipage}%
\hfill
\begin{minipage}[t]{0.29\textwidth}
\strut\vspace*{-\baselineskip}\newline\centering
    \setlength{\tabcolsep}{4pt}
\resizebox{0.95\columnwidth}{!}{
\renewcommand\arraystretch{1.05}
\begin{tabular}{lc|ccc}
\toprule
\textbf{Model}                        & \textbf{}              & \textbf{F1}       & \textbf{Acc.}      & \textbf{BLEU}     \\ \midrule
CoDial                                &                        & 58.5     & 60.1     & \textbf{45.2}    \\ \midrule
        \rowcolor{headercolor}\multicolumn{5}{l}{\textit{Refinement Instructions (RI)}} \\
~~~$-$ RI 3                           &                        & 56.1     & 58.0       & 41.6    \\
~~~$-$ RI 3 \& 2                      &                        & 55.9     & 57.6    & 41.7    \\
~~~$-$ RI 3 \& 2 \& 1                 &                        & 56.1     & 57.3     & 38.4    \\ 

\midrule
        \rowcolor{headercolor}\multicolumn{5}{l}{\textit{Generative Approach}} \\
~~~$-$ NAP                           &                        & 47.4     & 47.0       & 25.8    \\
~~~$-$ NAP \& DST                          &                        & 42.7     & 43.0       & 23.8    \\ \midrule
        \rowcolor{headercolor}\multicolumn{5}{l}{\textit{Oracle Vote}} \\
~~~DST: 4o-mini $+$ DSV3           &                        & 60.2    & 61.6     & 46.8    \\ \midrule
        \rowcolor{headercolor}\multicolumn{5}{l}{\textit{Silver Label DST}} \\
~~~$+$ STARv2 States              &                        & \textbf{60.7} & \textbf{62.9} & 44.3    \\ \bottomrule
\end{tabular}
}
\caption{Ablations on the STAR dataset.}
    \label{tab:star-abl}
\end{minipage}%
\hfill
\begin{minipage}[t]{0.35\textwidth}
\strut\vspace*{-\baselineskip}\newline    \centering
    \setlength{\tabcolsep}{2pt}
    \resizebox{0.98\columnwidth}{!}{
    \renewcommand\arraystretch{1.05}
    \begin{tabular}{p{3.1cm} | c c c c c}
        \toprule
        \textbf{Model}                             & \textbf{JGA} & \textbf{Inform} & \textbf{Success} & \textbf{BLEU} & \textbf{Combined} \\
\midrule
                \rowcolor{headercolor}        \multicolumn{6}{l}{\textit{Predicted Belief State}} \\
IG-TOD (few-shot)        & 27           & -               & 44               & 6.8           & -                 \\
CoDial \textsc{single}*           & 46.2         & 91.5            & 77.6             & 3.2           & 87.7             \\

CoDial                         & 28.4        & 76.6              & 54.6            & 3.5         & 69.1          
\\\midrule
\rowcolor{headercolor}        \multicolumn{6}{l}{\textit{Oracle Belief State}} \\
IG-TOD (few-shot) & -            & -               & 68               & 6.8           & -                 \\
CoDial \textsc{single}*     & -        & 94.6            & 90.6             & 3.5          & 96.1             \\
CoDial            & -         & 93.1            & 75.3             & 4.0          & 88.2               \\  
\midrule
\rowcolor{headercolor}        \multicolumn{6}{l}{\textit{DST Prompt Optimization}} \\
CoDial \textsc{auto}         & 31.1         & 72.3         & 57.2            & 3.6          & 68.3              \\                   
CoDial \textsc{manual}         & 28.5         & 80.4         & 57.9            & 3.6          & 72.7              \\                   
        \bottomrule
    \end{tabular}}
    \caption{Ablations on MultiWOZ. Settings with an asterisk (*) are not directly comparable due to a simpler task setup.}
    \label{tab:mwz-abl}
\end{minipage}
\end{table*}

\paragraph{Action Prediction and API Calls}
We find that NeMo Guardrails' intent detection performs strongly, achieving an F1 score of 96.3 on global actions (\Cref{tab:llm-gen-perf}). Additionally, we observe that STAR's API calling precision—measured as the ratio of correct API calls to the total number of API calls—stands at 74.9.
\Cref{tab:llm-gen-perf} also summarizes the performance of the actions that are generated by $\llma$ (i.e., when NAP component does not generate an output). LLM-generated actions account for $25\%$ of all predicted actions, with $70\%$ of them belonging to three fallback actions: \texttt{goodbye}, \texttt{out\_of\_scope}, and \texttt{anything\_else}. Excluding fallbacks, LLM-generated actions only account for $9.2\%$ of predictions, indicating that our NAP logic is generally effective at generating outputs based on the predicted state. Since fallback actions are a simple 3-way classification, we would expect high performance. However, $\llma$ achieves an F1 score of only $51.4$. We attribute this to the lack of an explicit schema for fallback actions in the STAR dataset, leading to inconsistencies in wizard annotations. Additionally, we observe a significant performance drop from fallback to non-fallback actions in both F1 ($51.4 \rightarrow 38.7$) and accuracy. This suggests that despite having an explicit schema, LLMs struggle to capture the more complex logic needed to predict non-fallback actions.
Our findings align with \citet{BuildingGuardrailsLarge-dong-2024}, reinforcing the need for a neuro-symbolic approach. 

\begin{figure}[t]
  \centering
  \includegraphics[width=.98\linewidth]{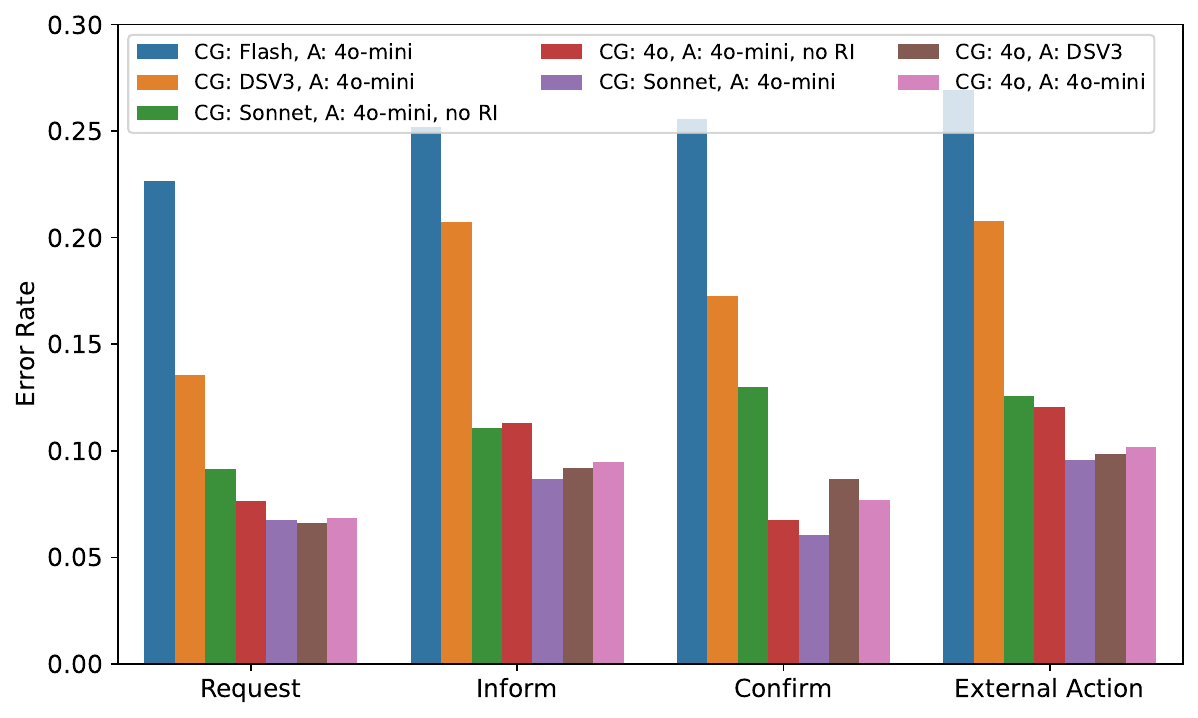}

   \caption{Error rate comparison of agents' predicted state on the STAR dataset across different node types, coloured by ($\llmcg$, $\llma$) pairs.}
   \label{fig:state-estimation}
   
\end{figure}

\paragraph{State Prediction}

\Cref{fig:state-estimation} shows the error rate of the predicted conversation state across different node types for each model. To approximate the error, we compare the model's predicted state with the estimated ground-truth state (i.e., the wizard’s state), as described in \Cref{ap:star}. We find that the error rate generally inversely correlates with the overall performance in \Cref{tab:star-only}; higher-performing models
tend to exhibit lower state prediction error.

\paragraph{Single- vs. Multi-Domain Performance}

Most of the MultiWOZ test set consists of multi-domain conversations, where a user may, for example, book both a taxi and a restaurant in the same dialogue. Since CoDial is designed for single-domain interactions, we report its performance on single-domain dialogues in \Cref{tab:mwz-abl}, where it performs well. However, with our naive oracle domain setting, CoDial performance drops significantly. This is likely due to compounded errors from DST to NAP, which we analyze further in \Cref{sec:ablations}. The DST performance decrease is also observed in other baselines, as shown in \Cref{tab:mwz-single-vs-multi}.

\vspace{-2mm}
\subsection{Ablation Studies}
\vspace{-0.5mm}
\label{sec:ablations}
\paragraph{Oracle DST Performance}

To assess the impact of DST error, we evaluate CoDial under an Oracle setting. Since STAR does not provide gold DST labels, we simulate an oracle setting by using the silver annotations from STARv2 (\Cref{tab:star-abl}). This results in a performance gain of $+2.2$ F1 and $+2.8$ accuracy. We do the same for MultiWOZ, where we use the gold belief states, which leads to a substantial performance improvement (\Cref{tab:mwz-abl}).
These findings suggest that investigating more advanced DST approaches, such as inference-time scaling explored in \Cref{sec:star-mwz}, could be a promising direction to improve performance.
We also experiment briefly with prompt optimization for lower cost DST improvements, described below.
\paragraph{Code Optimization} 

We use LLMs to perform iterative code refinement and automatic prompt optimization for the DST prompts. Refining the code with RIs consistently enhances CoDial’s performance, demonstrating the benefits of integrating user feedback into the generation process. After prompting the LLM to iteratively refine its outputs, CoDial achieves better accuracy and fluency (compared to CoDial $-$ \textsc{ri} in \Cref{tab:star-only}). 
We also conduct an ablation study to examine the effect of the individual RIs, summarized in \Cref{tab:star-abl}. Although all RIs are beneficial, most of the performance improvements can be attributed to the third RI, which refines the conditional logic of request nodes.

After observing the results of the oracle DST setting, we also apply prompt optimization to improve DST accuracy. As shown in \Cref{tab:mwz-abl}, automatic prompt optimization yields only marginal gains across metrics, with the exception of Inform, suggesting that automatic DST improvement remains a non-trivial challenge. 
To explore the impact of human feedback, we also experiment with manual prompt optimization (\Cref{ap:mwz}), making minor edits to the prompts for the ``attraction'' domain. This results in consistent improvements across all metrics, reinforcing that human-crafted prompts can still outperform automatic optimization.

\vspace{-1mm}
\paragraph{Generative Approach}

To better understand the effectiveness of the proposed CoDial architecture, we experiment with a setting in which the NAP component is removed and all actions are predicted in a fully generative manner by the $\llma$, similar to \citet{SGPTODBuilding-zhang-2023}. We prompt the $\llma$ with simplified dialogue flows following their work and include DST predictions in the prompt. This results in a substantial drop in performance, as shown in \Cref{tab:star-abl}, highlighting the importance of our NAP approach. We further ablate the model by removing the DST component. The performance is a smaller reduction than NAP alone. This suggests synergy between NAP and DST; the system performs best when both are strong.

\vspace{-1mm}
\paragraph{Usage and Cost.} We perform a cost analysis of CoDial (\texttt{4o, Qwen}) on STAR, summarized in \Cref{tab:cost} and depicted in \Cref{fig:cost}. The average cost is \$1.63 per 1,000 dialogues and \$0.27 per 1,000 turns, ranging from \$0.16 to \$0.38 across the 24 tasks and scaling with task complexity. Crucially, these inference costs should be considered alongside the data costs avoided: unlike SOLOIST and MARS, CoDial requires no annotated training data, making it well-suited for domains where human annotation is scarce.

\vspace{-1mm}
\subsection{Human Study}

\vspace{-0.75mm}
In addition to being interpretable by design via explicit guardrail representations, we conduct a human study to quantitatively evaluate CoDial’s interpretability. We recruited three non-author participants with no prior exposure to CoDial or Colang and compared CoDial against a prior work, SAM, on 50 randomly sampled conversation turns. Participants provided preference judgments on response quality, and rated ease of understanding using a 5-point Likert scale. As shown in \Cref{tab:human-interpretability}, CoDial is preferred in $\approx$69--71\% of cases, while SAM is preferred in fewer than 7\%. CoDial also achieves a higher average score, with a mean Likert increase of 1.8 points over SAM ($p<0.001$, one-tailed paired $t$-test). We also showed the annotators two examples of code generated with \codials, and annotators were moderately confident they could understand the code after seeing 50 conversation samples. Refer to \cref{ap:human-study} for more details.

\begin{table}[t]
\centering
\setlength{\tabcolsep}{8pt}
\renewcommand\arraystretch{1.05}
\resizebox{0.7\linewidth}{!}{
\begin{tabular}{l|ccc}
\toprule
\textbf{Criterion} & \textbf{CoDial} & \textbf{SAM} & \textbf{Tie} \\
\midrule
\rowcolor{headercolor}
\multicolumn{4}{l}{\textit{Human Preference (\%)}} \\
Conversation History & 70.7 & 5.4 & 23.9 \\
Dialogue Flow & 68.7 & 6.1 & 25.2 \\
\midrule
\rowcolor{headercolor}
\multicolumn{4}{l}{\textit{Ease of Understanding}} \\
Likert 1--5 & 4.27 {\scriptsize$\pm$ 0.87} & 2.46 {\scriptsize$\pm$ 1.16} & - \\
\bottomrule
\end{tabular}}
\caption{Human evaluation of interpretability.}
\label{tab:human-interpretability}
\end{table}

%% file: sections/6_conclusion.tex
\vspace{-1.mm}
\section{Conclusion}
\vspace{-0.75mm}

In this work, we introduced CoDial, a novel framework for building interpretable TOD systems by grounding structured dialogue flows to programmatic guardrails. CoDial introduces CHIEF, a heterogeneous graph representation of dialogue flows, and employs LLM-based code generation to automatically convert dialogue flows into executable guardrail specifications (e.g. NVIDIA’s Colang), enabling zero-shot creation of interpretable TOD systems. Through manual and LLM‐aided refinements, CoDial supports rapid incorporation of user feedback, further enhancing the generated code. Our empirical findings support CoDial’s effectiveness, achieving SOTA performance on STAR and competitive results on MultiWOZ in a strict zero-shot setting.

\newpage

\section*{Limitations}

While CoDial offers an interpretable and modifiable approach to TOD systems, it has certain limitations. 
First, scalability remains a challenge. For large and complex dialogue flows, CoDial re-queries all slots every turn, which may increase latency and computational cost. In general, improving DST performance and efficiency remains a potential direction for future work.

Second, CoDial is less effective for multi-domain dialogues, as it operates on a single dialogue flow at a time. Several directions could extend CoDial to handle domain transitions. One approach is model calibration: if the predicted response confidence under the current domain falls below a threshold, this could trigger a domain switch. However, low confidence may also arise when users deviate from the expected conversation structure for unrelated reasons, requiring mechanisms to disambiguate these sources of uncertainty. Alternatively, Hidden Markov Models could directly model transition probabilities based on user input (e.g., detecting phrases like "Can I also…"), though this requires observing domain transition patterns and may limit generalization to unseen domain pairs. We leave the design of such a system to future work.

Moreover, developing measurable metrics for user accessibility, a central motivation of this work, remains an open direction. An ideal study would evaluate the effort required for users to represent their knowledge as a task schema (e.g., CHIEF representation in CoDial) and compare it across approaches. While CoDial abstracts away manual programming, certain applications may still require some familiarity with LLM guardrails and Colang for effective modification. CoDial mitigates this through textual refinement interfaces (RIs), though their adequacy ultimately depends on the specific use case.

\subsection*{Ethics Statement}

This work adheres to ethical research practices by ensuring that all models, codebases, and datasets used comply with their respective licenses and terms of use. The STAR and MultiWOZ datasets employed in our experiments do not contain personally identifiable information or offensive content.

As with any system leveraging LLMs, CoDial inherits potential risks related to bias and factually incorrect outputs. However, our framework mitigates these risks by enforcing structured dialogue flows, guardrailing based on user intent, and template-based responses, reducing the likelihood of hallucinated or biased content.
Future work may integrate NeMo Guardrails’ input and output rails to filter inappropriate inputs and outputs, enhancing system safety. Since our focus is on structured dialogue flows, we leave this for future exploration.

%% file: sections/appendix.tex
\section{Appendix}

\subsection{Detailed Related Work}\label{ap:rw}

\paragraph{Task-Oriented Dialogue}
Building generalizable conversational systems is challenging due to the complexity of human conversations, particularly when domain expertise is involved \citep{chen2017survey}, leading to a focus on task-oriented systems for specific domains \citep{Doyoufollow-jacqmin-2022}. While LLMs have demonstrated impressive capability in a wide variety of domains, they struggled with TOD and fell behind if not used properly \cite{AreLargeLanguage-hudecek-2023}. Some research \citep{SGPTODBuilding-zhang-2023, SchemaGuidedParadigm-mehri-2021} has used a neural schema-guided approach to generalize TOD systems to unseen tasks without interpretability. AnyTOD \citep{AnyTODProgrammableTask-zhao-2023} provided an interpretable neuro-symbolic approach by viewing task schema as a manually-written policy program that controls the dialogue flow. However, beyond the manual coding requirement, AnyTOD also relied on extensive training with highly similar task schemas. As a result, it suffered substantial performance drops when transferred to even slightly different task structures, revealing limited generalizability to unseen tasks.
Recent unsupervised methods aim to automatically induce dialogue flows or schemas from raw conversations, reducing manual design effort and improving scalability \citep{UnsupervisedExtractionDialogue-sreedhar-2024, UnsupervisedLearningHierarchical-lu-2022, WorkflowDiscoveryDialogues-hattami-2023}. These works are orthogonal to our work---as mentioned in \Cref{sec:df-framework}, we demonstrate that we can take input schemas and enrich them further with our heterogeneous graph representation.

\paragraph{Guardrails} CoDial employs guardrails to steer LLMs behaviour. Guardrailing aims to enforce human-imposed constraints to control LLMs in the inference time \citep{BuildingGuardrailsLarge-dong-2024,NeMoGuardrailsToolkit-rebedea-2023,guardrailsai}. While originating from AI safety, we argue that they can generally be used to define desired behaviour to constrain. Although traditional dialogue management systems, like Google Dialogflow\footnote{\url{https://dialogflow.cloud.google.com}}, allow rigid modelling of dialogue states, they often lack the flexibility to define complex task logic, and it is difficult for a user to further enhance the system.
NVIDIA NeMo-Guardrails \citep{NeMoGuardrailsToolkit-rebedea-2023} is a toolkit that adds programmable guardrails to LLM-based conversational applications without fine-tuning. NeMo-Guardrails employs Colang \cite{nvidia2024nemo}, a programming language, to establish highly flexible conversational flows and guide LLMs within them. 
More broadly, a range of guardrailing languages and frameworks \citep{dong2024safeguarding} can serve a similar role in CoDial, including programmatic approaches such as Guidance AI\footnote{\url{https://github.com/guidance-ai/guidance}} and LMQL\footnote{\url{https://lmql.ai/}}, as well as specification-based tools like Guardrails AI, which uses XML-style RAIL definitions \citep{guardrailsai}. We select Colang specifically for its user intent detection and its support for natural language descriptions.
\citet{BuildingGuardrailsLarge-dong-2024} suggested using neuro-symbolic approaches to guardrail LLMs, where a neural agent (e.g., an LLM) can deal with frequently seen cases, and a symbolic agent can embed human-like cognition through structured knowledge for the rare cases.

\begin{figure*}[ht]
  \centering
  \begin{subfigure}[c]{0.48\textwidth}
    \centering
  \includegraphics[width=0.99\linewidth]{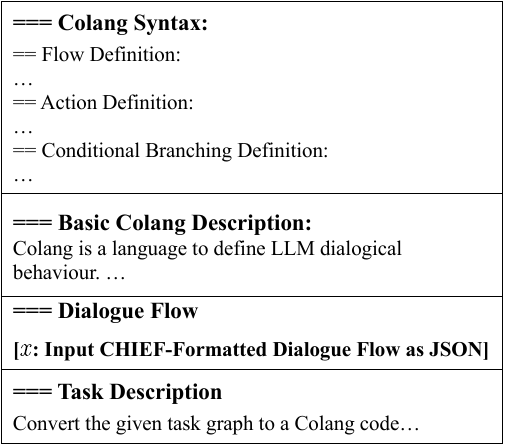}

   \caption{$\promptcg$ for \codialf
   }
   \label{fig:prompt-overview2}
  \end{subfigure}
  \hfill
  \begin{subfigure}[c]{0.48\textwidth}
    \centering
  \includegraphics[width=0.99\linewidth]{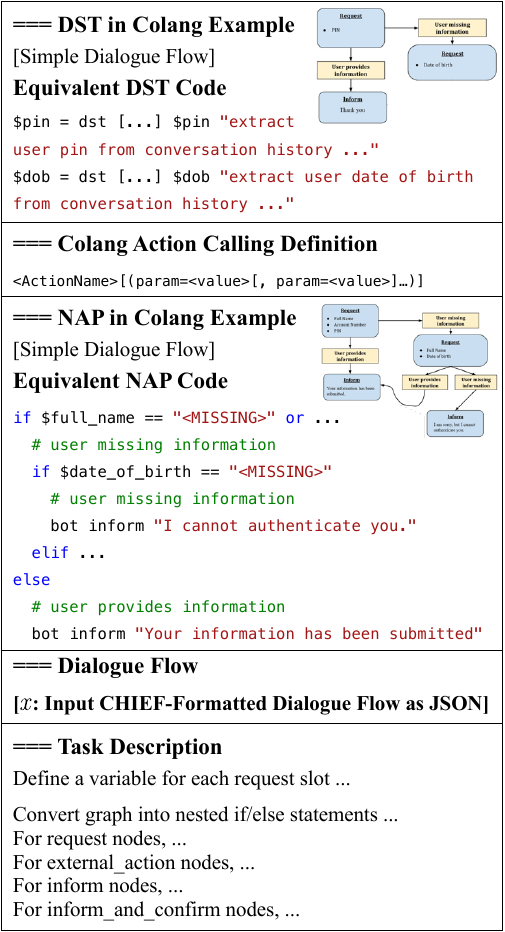}

   \caption{$\promptcg$ for \codials
   }
   \label{fig:prompt-overview}
  \end{subfigure}
  \caption{An overview of $\promptcg(x)$, where a dialogue flow $x$ is wrapped with system prompt template.}
  \label{fig:prompts}
\end{figure*}

\begin{table*}[ht]
    \centering
    \resizebox{0.75\linewidth}{!}{
    \begin{tabular}{l|l|p{10cm}}
        \toprule
        \textbf{ID}   & \textbf{Description}    & \textbf{Instruction} \\
        \midrule
\textbf{RI 1} & Revise if statements    & Revise the `if`s to exactly reflect the nodes. Comment each `if` to specify the corresponding node ID. Make sure the generated `if` statement and its body reflect the instructions for that node type. \\ \hline
\textbf{RI 2} & Fix dst dependent vars  & Fix dst's first input parameter. It should reflect which variables should be invalidated when the corresponding slot is updated.                                                                        \\ \hline
\textbf{RI 3} & Fix request node checks & Fix `if` checks for request nodes. Comment their rule, if available. The `if` should reflect the rule for each node.                                                                                    \\
        \bottomrule
    \end{tabular}}
    \caption{Instructions for Code Refinement}
    \label{tab:ref-ins}
\end{table*}

\paragraph{Colang}
Colang is an event-driven interaction modelling language designed for adding guardrails to LLM-powered conversational systems. Colang models the interaction between an application and an LLM as a stream of events—including user utterances, LLM-generated responses, action triggers, and guardrail activations. The language centers on three core abstractions: flows (sequences of messages and events with branching logic), events (structured representations of what happens during conversation), and actions (custom functions for external operations). The Colang runtime recognizes and enforces patterns within the event stream, enabling developers to specify conversational constraints through flow definitions that match against canonical message forms and context variables. This event-driven architecture provides a flexible foundation for controlling LLM behaviour throughout complex multi-turn text-based interactions.

Two features are particularly relevant to our method. \texttt{continuation on unhandled user intent} invokes the LLM when a user intent does not match any predefined flow, to determine a suitable continuation. Natural Language Descriptions (NLDs) are natural-language specifications evaluated by the LLM at runtime to generate or extract context-dependent values (e.g., summaries, classifications, or decisions) that are then consumed by flows, enabling guarded LLM reasoning to be embedded within an otherwise deterministic interaction structure.

\paragraph{Code Generation and Prompt Optimization} We use code generation strategies to convert structured graphs into programmatic guardrails. Code generation has made remarkable progress with the introduction of LLMs \cite{NEURIPS2022_8636419d}. Although there are still challenges such as logical consistency and hallucinations \cite{liu2024your}. LLMs are proficient when in-context examples, documentations, or plans are provided~\cite{jiang2024self}. There are many emerging methods to further optimize LLM generations (e.g., self-reflection, where LLMs are requested to update their own response), which have been shown to reduce hallucinations and improve problem solving \cite{ji2023towards}. There has been research to improve output by rewriting the input prompt, referred to as prompt optimization \cite{yang2023large, yuksekgonul2024textgrad}. 

\subsection{Details on CHIEF and GCG} \label{ap:cg}

\subsubsection{CHIEF} \label{ap:det-chief}

Below, we discuss the main node types and actions in CHIEF.

\paragraph{Request}  
The request nodes define the variables, hereby referred to as slots, that CoDial tracks throughout the conversation (e.g. \textit{the departure location in a taxi booking task}). When a conversation reaches this node, the system will request information specified by the slots. Each slot is assigned a data type (e.g. categorical) and accompanied by a few example values. Additionally, CHIEF includes a free-form \texttt{rule} property to define the conditions under which a slot should be requested (e.g. in a taxi booking scenario, providing either a departure or arrival time is sufficient for booking). Since we leverage LLMs to build the TOD system, textual extensions can be easily incorporated.
\paragraph{External Action}
This node specifies a call to an external function within a dialogue flow. External actions enable the designer to execute complex logics through programming functions, interact with APIs, or invoke an LLM.

\paragraph{Inform (and Confirm)}
This node defines a template for providing information to the user (e.g. \textit{Your taxi is booked with reference number [ref\_no]}). The confirmation variant additionally allows the agent to ask a follow-up question (e.g. \textit{Do you confirm the booking?}) and follow the appropriate predefined dialogue path based on the user’s response.

\paragraph{Global and Fallback Actions}
In addition to nodes, CHIEF supports representing global and fallback actions that are not tied to particular dialogue steps. Global actions can be triggered at any point in the dialogue flow (e.g. responding to a greeting). We also define fallback actions, general responses used when no other action is selected (e.g. \textit{Sorry, I can't help with that}).

\begin{figure*}[ht]
  \centering
  \begin{subfigure}[c]{0.48\textwidth}
    \centering
    \includegraphics[width=\linewidth]{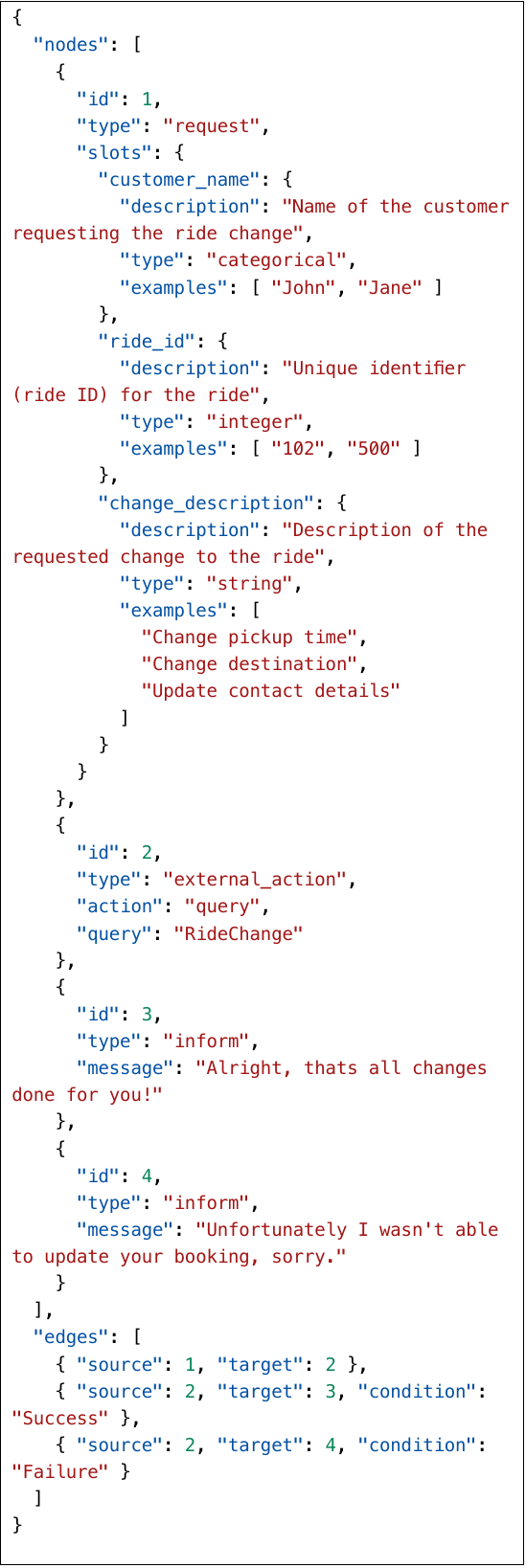}
    \caption{Converted JSON representation for STAR \textit{Ride Change} task}
    \label{fig:example-json}
  \end{subfigure}
  \hfill
  \begin{subfigure}[c]{0.48\textwidth}
    \centering
    \includegraphics[width=\linewidth]{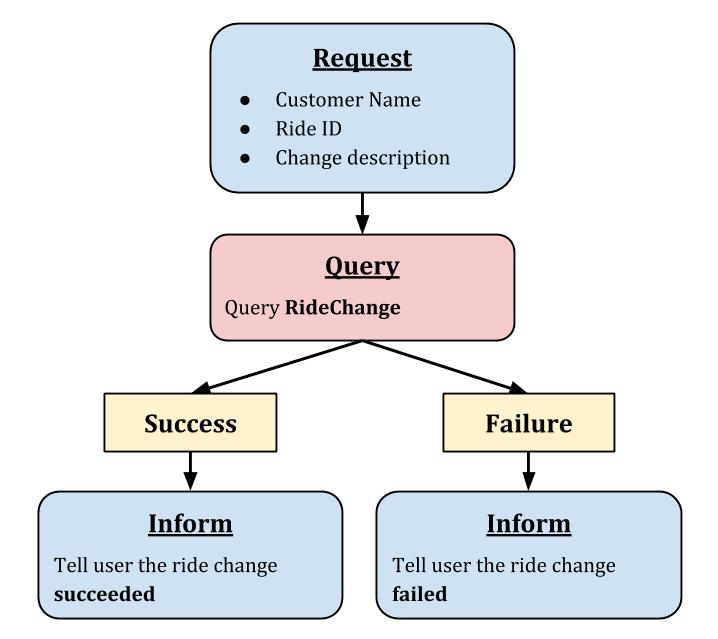}
    \caption{STAR \textit{Ride Change} task schema}
    \label{fig:example-df}
  \end{subfigure}
  \caption{Example of STAR task schema and converted JSON object}
  \label{fig:example-comparison}
\end{figure*}

We represent the graphs defined by CHIEF (\Cref{sec:df-framework}) as text in JSON format. The JSON representation consists of a list of nodes and a list of edges. The node list defines the dialogue flow nodes, specifying their types and assigning each a unique identifier (node ID). The edge list specifies the connections between nodes using their IDs (e.g., \Cref{fig:example-json}). The JSON nodes, global and fallback actions, and functional specifications for function calls are translated into Colang code with our automatic code generation pipeline. The external action node functions, referred to as Actions in Colang, are implemented in Python.

\subsubsection{GCG} \label{ap:det-cg}

\paragraph{Dialogue State Tracking (DST)}
Since updating a slot may affect the state (e.g. in a search task, modifying the search criteria requires re-executing the search), CoDial needs to identify the helper variables that need to be invalidated when each slot is updated. We instruct $\llmcg$ to list the helper variables of nodes that are reachable from the updated slot in the graph (i.e., nodes that are direct or indirect children of the slot's request node). These variables are then reset to null or false, depending on their type, when the slot is updated during execution.

\paragraph{Post-processing}  
Following code generation by the LLM, we apply rule-based post-processing to ensure proper execution. This includes adding helper flows (Colang’s equivalent of functions) to support algorithm execution, enabling the loading of the STAR API function, and injecting additional code for evaluation purposes.

\paragraph{Helper Variables}  

The \codials{} algorithm designed in Colang (\Cref{alg:main}) determines whether a request node should be executed (i.e., prompt the user for information) by checking the values of its associated slots. To track the state of other node types, we instruct $\llmcg$ to define helper variables following a structured naming pattern, where \texttt{<id>} represents the corresponding node's ID:  

\begin{itemize}  
    \item \texttt{action\_<id>}: Stores the return value of external actions.  
    \item \texttt{inform\_<id>}: Indicates whether the node has been executed and the user has been informed.  
    \item \texttt{answered\_<id>}: For inform and confirm nodes, stores the user's response.  
\end{itemize}

\subsection{STAR Implementation Details} \label{ap:star}
\paragraph{API Calling}
While not the primary focus of this paper, we use prompting to generate Colang's Python action code for calling STAR's API and processing its outputs automatically, rather than directly feeding ground-truth API responses as input as done in other works. Every piece of code in our pipeline is automatically generated.
Since STAR's API returns randomized outputs, we return the ground-truth API response object when it is available for the exact same turn, instead of the random sampling response.
\paragraph{Dialogue Flows}
We convert the STAR task schemas, originally provided as images, into CHIEF representation described in \Cref{sec:df-framework}. We use one-shot prompting with GPT-4o to convert pictures into JSON. We convert yellow nodes in pictures into conditions for edges. However, we observed that GPT-4o occasionally misassigns edge connections, requiring manual corrections. Additionally, we enrich the JSON representations by adding more context, such as example values for each slot. We also define \texttt{hello} action as the only global action and \texttt{goodbye}, \texttt{out\_of\_scope}, and \texttt{anything\_else} as fallback actions for all tasks.

To better align the dialogue flows with the actual collected dialogues, we introduce minor modifications, such as adding the \verb|inform_nothing_found| action for search tasks. We also identified small inconsistencies between the provided API schema and its implementation. To address this, we refine the API definitions and modify the sampling logic to prevent errors when no results match the given constraints. We will release these improvements, aiming to support future research.

\paragraph{Wizard State Approximation} For \codials{} evaluation, since we are working with offline conversations (i.e., the user is not interacting with the actual TOD system), we approximate the wizard's state at the end of each turn and adjust the program's state accordingly. This helps prevent the program's state from deviating from the ground-truth conversation. To achieve this, we first find the node in dialogue flow that the ground-truth conversation was in by mapping the ground-truth action label, if available, to a node in the dialogue flow. We manually create this mapping from action labels to the dialogue flow nodes. Next, we use depth-first search to trace the path from the start of the dialogue flow to the current conversation node. Finally, we adjust each state variable based on whether the corresponding node is part of the current conversation pathway, as described in \Cref{alg:state-correction}.

\begin{algorithm}[t]
\caption{Wizard state approximation}
\label{alg:state-correction}
\begin{algorithmic}[1]
\Require Variable $v$, Graph $G$, Ground-truth action $a_{gt}$, Mapping $\phi$
\Ensure Approximated value or \textsc{null}
\State $n_{tgt} \gets \phi(a_{gt})$
\State $n_v \gets v.node$
\State $P \gets$ \Call{DFSPath}{$G, G.start, n_{tgt}$}
\If{$n_v \notin P$}
    \State \Return \textsc{null}
\EndIf
\For{each $e \in P$}
    \If{$e.target = n_v$}
        \State $e_v \gets e$
        \State \textbf{break}
    \EndIf
\EndFor
\State \Return \Call{ApproxValue}{$e_v.condition, v$}
\end{algorithmic}
\end{algorithm}

\paragraph{Prompt Context}  
During evaluation, we incorporate the textual guidelines provided to wizards into $\llma$'s context. This additional context helps the LLM infer some details, such as the time or location of the conversation. For example, a guideline might look like: \textit{Some facts you should be aware of: Right now, it is Tuesday, 12 PM.}  

\subsection{MultiWOZ Implementation Details}
\label{ap:mwz}
We preprocess MultiWOZ 2.2 using the code from \citet{FnCTOD-li-2024} to annotate each conversation turn with its active domains. For each turn $i$, we use the dialogue flow(s) of the corresponding domain(s) to predict the output and merge all turns at the end.
\paragraph{Manually Crafted Dialogue Flows}Unlike STAR, MultiWOZ does not provide explicit dialogue flows for each domain, nor do its conversations adhere to a specific flow. To address this, we manually construct simple dialogue flows by analyzing a few example dialogues from each domain. We will release these crafted MultiWOZ dialogue flows. Additionally, for evaluation, we modify the prompts and instruct the LLM to generate delexicalized texts.\footnote{Refer to \citet{ShadesBLEUFlavours-nekvinda-2021} for more details.}

\paragraph{Naive Multi-domain}
Rather than adding a separate domain detection step, we use the gold labels for the active domains at each conversation turn and directly apply the corresponding dialogue flows. We preprocess MultiWOZ 2.2 using the code from \citet{FnCTOD-li-2024} to annotate each turn with its active domains. Since evaluation is offline, we separate turns in a conversation by domain, simulate the conversation with prior history, and use the corresponding Colang program(s). Finally, we merge all turns and treat slots from all domains as a single set, accumulating DST predictions during evaluation.

\begin{algorithm}[t]
\caption{Our prompt optimization algorithm. We randomly sample a training and validation set of size 20 and 50 for every DST slot, respectively.}\label{alg:textgrad}
\begin{algorithmic}[1]
\Require Training set $\mathcal{D}_{\text{train}}$, Validation set $\mathcal{D}_{\text{val}}$, Instruction $I$, Agent $\llma$, Optimizer LLM $M$, Batch size $B$
\State $\hat{Y}_{\text{val}} \gets $ \Call{DST}{$\mathcal{D}_{\text{val}}.H, \llma, I$}
\State Initialize $S \gets$ \Call{ComputeScore}{$\hat{Y}_{\text{val}}, \mathcal{D}_{\text{val}}.Y$}
\State $I_{\text{best}} \gets$ $I$
\State Divide $\mathcal{D}_{\text{train}}$ into batches $\mathcal{B}_1, \ldots, \mathcal{B}_n$ of size $B$
\For{each batch $\mathcal{B}$ in $\mathcal{D}_{\text{train}}$}
    \State $(H, Y) \gets \mathcal{B}$
    \State $\hat{Y} \gets$ \Call{DST}{H, $\llma$, $I_{\text{best}}$}
    \State $I \gets$ \Call{$M$.Rewrite}{$H, \hat{Y}, Y, I$}
    
    \State $\hat{Y}_{\text{val}} \gets $ \Call{DST}{$\mathcal{D}_{\text{val}}.H, \llma, I$}
    \State $S \gets$ \Call{ComputeScore}{$\hat{Y}_{\text{val}}, \mathcal{D}_{\text{val}}.Y$}
    
    \If{$S > S_{\text{best}}$}
        \State $S_{\text{best}} \gets S$
        \State $I_{\text{best}} \gets I$
    \EndIf
\EndFor
\State \Return $I_{\text{best}}, S_{\text{best}}$
\end{algorithmic}
\end{algorithm}

\paragraph{DST Prompt Optimization}
The NAP component's performance is largely dependent on DST, as the next action is determined by the values known to the dialogue system (Equation \ref{eq:nap}). However, we found in preliminary experiments that the DST performance can be poor with original $P^{(s)}$ prompts, generated by general guidelines outlined in $\promptcg$. To this end, we further refine $P^{(s)}$ with automatic prompt optimization.

Our optimization algorithm is summarized in \Cref{alg:textgrad}. For each DST variable $v^{(s)}_j$, we randomly sample two mutually exclusive sets of conversation turns to serve as training and validation sets. The training examples are divided into batches of 5, and each batch is used to guide the optimizer GPT-4o model to rewrite the instruction $p^{(s)}_j$, resulting in a candidate prompt. If the revised instruction improves performance on the validation set, it is retained; otherwise, the original is kept, ensuring that modifications are only accepted when they lead to measurable improvements.

In addition, we manually refine the prompts for the worst-performing domain, ``attraction.'' The edits include defining what an ``attraction'' is by listing all possible types, and propagating the predicted \texttt{type} value to other slot instructions to maintain consistency. We leave further investigation of this technique—passing key slot predictions across instructions within a domain—as future work.

\subsection{Experimental Details} \label{ap:exp}

We use GPT-4o\footnote{\url{https://openai.com/index/hello-gpt-4o/}}, Claude 3.5 Sonnet\footnote{\url{https://www.anthropic.com/news/claude-3-5-sonnet}}, Gemini 2.0 Flash\footnote{\url{https://developers.googleblog.com/en/gemini-2-family-expands/}}, and DeepSeek V3 (DSV3) \citep{DeepSeekV3Technical-deepseek-ai-2024} as $\llmcg$, and GPT-4o-mini, GPT-5, Qwen-3-30B-A3B Instruct \citep{yang2025qwen3}, and DSV3 as $\llma$. We access OpenAI models through OpenAI and other models through OpenRouter\footnote{\url{https://openrouter.ai/}} API.

If a generated program contains syntax or runtime errors, we regenerate the code to obtain a functional version. The only exception is Gemini 2.0 Flash, which struggles with calling our defined Colang helper flows. Since this issue is minor, we manually correct the syntax to assess the model’s ability to generate programmatic logic for dialogue flows. \Cref{tab:code-gen-success} reports statistics on GPT-4o code generation success rates. Examples of failure modes include the use of unsupported syntax constructs, such as \texttt{end} tags for \texttt{if} statements, or unsupported operations such as inline string formatting with variable indexing.

\begin{table}[t]
\centering
    \renewcommand\arraystretch{1.05}
    \resizebox{0.65\linewidth}{!}{
\begin{tabular}{lc}
\toprule
\textbf{Metric} & \textbf{GPT-4o Success Rate} \\
\midrule
Per-task min & 2 (50\%) \\
Per-task max & 4 (100\%) \\
Average      & 3.45 (86\%) \\
\bottomrule
\end{tabular}}
\caption{Colang code generation success rate of GPT-4o over 24 STAR task schemas, with 4 trials per schema (96 generations total). A generation is considered successful if it passes Colang syntax checking.}
\label{tab:code-gen-success}
\end{table}

\paragraph{NeMo-Guardrails}
To implement the Colang guardrails, we use a fork from NeMo-Guardrails version 0.11, modified to inject our evaluator class\footnote{The modified NeMo-Guardrails version that we used for the experiments is available at \url{https://github.com/radinshayanfar/NeMo-Guardrails/tree/paper}.}. We use this class to evaluate on the ground truth user-wizard history, instead of the history of user-bot conversation, similar to \citet{ShadesBLEUFlavours-nekvinda-2021}.

\begin{figure}[ht]
  \centering
  \includegraphics[width=\linewidth]{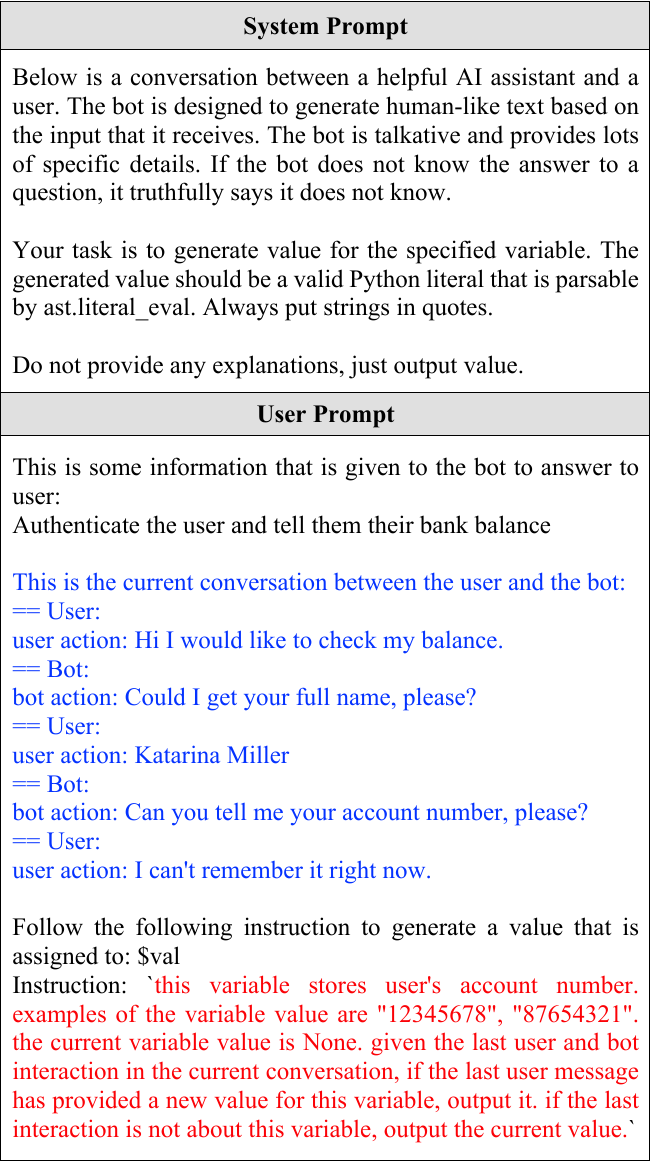}

   \caption{Example of the modified NeMo \texttt{value\_from\_instruction} action prompt, which is used for DST. $ \color{blue} h_{2i-1}$ and $ \color{red} p^{(s)}_j$ are provided in each prompt to generate a value for that slot.}
   \label{fig:dst-prompt}
\end{figure}

We modify NeMo's default \texttt{value\_from\_instruction} prompt structure to begin with a system message, followed by the entire conversation history and instructions combined into a single user message (\Cref{fig:dst-prompt}). During our initial experiments, we suspected that NeMo's original prompt structure—where each message in the conversation history was passed as a separate user or assistant message—hindered $\llma$'s ability to follow instructions effectively.  

Additionally, we refine the post-processing of this action. We found that $\llma$ was inconsistent in formatting return values, sometimes enclosing strings in quotation marks while omitting them for non-string types. To address this, we first check if both leading and trailing quotation marks are present and remove them if so. We then attempt to evaluate the return value as a Python literal. If this evaluation fails, we then enclose the value in quotation marks to ensure proper parsing as a string.

Moreover, we fixed an issue related to if-else statements in the Colang parser, which was later merged into the official NeMo repository\footnote{GitHub pull request at \url{https://github.com/NVIDIA/NeMo-Guardrails/pull/833}.
}.

\subsection{Detailed Baselines} \label{ap:baselines}

\begin{table}[t]
    \centering
    \setlength{\tabcolsep}{8pt}
    \renewcommand\arraystretch{1.05}
    \resizebox{0.95\linewidth}{!}{
    \begin{tabular}{l l | c c}
        \toprule
\textbf{Model}  &          & \textbf{F1}      & \textbf{Acc.}     \\        \midrule
        SGP-TOD \textsc{gpt3.5-e2e} &   & 53.5             & 53.2       \\
        SGP-TOD \textsc{gpt4o-mini-e2e} &   & 41.3             & 44.3       \\
        SGP-TOD \textsc{gpt4o-mini-e2e} Adapted &   & 40.3             & 43.8       \\
        \bottomrule
    \end{tabular}}
    \caption{Comparison of SGP-TOD baselines.} 
    \label{tab:sgp-tod}
\end{table}

\begin{figure}[ht]
  \centering
  \begin{subfigure}{\linewidth}
    \centering
    \includegraphics[width=\linewidth]{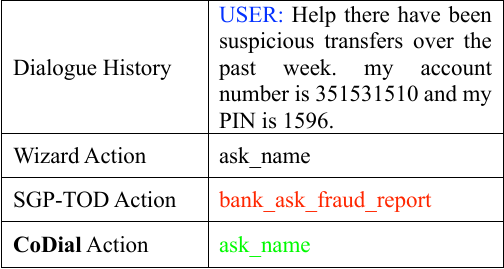}
    \caption{\textit{Bank Fraud Report} example dialogue. SGP-TOD fails to collect all necessary authentication details before requesting fraud report information, as its schema defines the next action after \texttt{user\_bank\_inform\_pin} as \texttt{bank\_ask\_fraud\_details}. In contrast, CoDial verifies that all required information is provided at each request node before proceeding, correctly identifying that the user's name is missing.}
  \end{subfigure}

  \vspace{0.5cm}
  \begin{subfigure}{\linewidth}
    \centering
    \includegraphics[width=\linewidth]{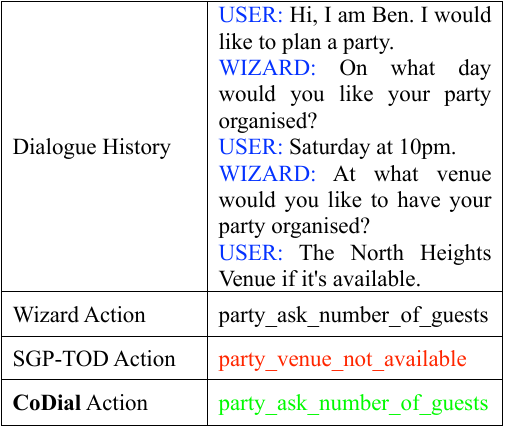}
    \caption{\textit{Party Plan} example dialogue. SGP-TOD produces an incorrect and uninterpretable prediction. In contrast, CoDial follows a programmatic logic aligned with the dialogue flow, ensuring interpretability.}
    \label{fig:sgp-tod-incorrect-b}
  \end{subfigure}

  \caption{Cherry-picked comparison of CoDial and SGP-TOD performance. We use GPT-4o-mini to reproduce SGP-TOD results.}
  \label{fig:sgp-tod-incorrect}
\end{figure}

\begin{figure*}[ht]
  \centering
  \includegraphics[width=.99\linewidth]{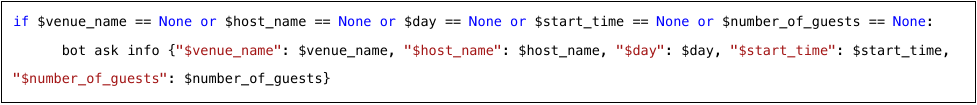}

   \caption{Example of code logic in CoDial that enables interpretability. A user can inspect runtime variables to trace the reasoning behind the outputs generated by the TOD system.}
   \label{fig:sgp-tod-incorrect-b-code}
\end{figure*}

\begin{itemize}
    \item \textit{SGP-TOD} \citep{SGPTODBuilding-zhang-2023} is a purely generative approach that uses two-stage prompting to track dialogue state and generate response. It employs graph-based dialogue flows to steer LLM actions, ensuring adherence to predefined task policies without requiring fine-tuning or training data.
    To ensure a fair comparison, we replicated their setup using the same newer $\llma$ model as ours (\Cref{tab:sgp-tod}). We ran their released code without modification, except for switching the API model to GPT-4o-mini. Surprisingly, performance dropped significantly. After contacting the authors, they advised adapting the prompt structure to the aligned LLMs—placing instructions in the system message and including examples and dialogue history in the user message. However, even with this adaptation, the performance did not match the results originally reported with GPT-3.5, suggesting that a generative approach could not be a trivial solution and requires careful prompt engineering. \Cref{fig:sgp-tod-incorrect} further illustrates differences between CoDial and SGP-TOD through two cherry-picked examples. Specifically, in \Cref{fig:sgp-tod-incorrect-b}, by analyzing the variable values in the runtime, a user can easily spot that the generated output stems from the code snippet in \Cref{fig:sgp-tod-incorrect-b-code}, where it asks for the next missing value, if any.
    \item \textit{BERT + Schema} and \textit{Schema Attention Model (SAM)} \citep{STARSchemaGuided-mosig-2020,SchemaGuidedParadigm-mehri-2021} incorporate task schemas by conditioning on the predefined schema graphs, enabling structured decision-making in TODs. SAM extends BERT + Schema approach with an improved schema representation and stronger attention mechanism, aligning dialogue history to the schema for more effective next-action prediction. Both models rely on fine-tuning to learn schema-based task policies and improve generalization across tasks.
    \item \textit{SOLOIST} \citep{SoloistBuildingTask-peng-2021} is a Transformer-based model that unifies different dialogue modules into a single neural framework, leveraging transfer learning and machine teaching for TOD systems. It grounds response generation in user goals and database/knowledge, enabling effective adaptation to new tasks through fine-tuning with minimal task-specific data.
    \item \textit{MARS} \citep{MarsModelingContext-sun-2023} is an end-to-end TOD system that models the relationship between dialogue context and belief/action state representations using contrastive learning. By employing pair-aware and group-aware contrastive strategies, Mars strengthens the modelling of relationships between dialogue context and semantic state representations during end-to-end dialogue training, improving dialogue state tracking and response generation.
    \item \textit{DARD} \citep{DARDMultiAgent-gupta-2024} is a multi-agent TOD system that delegates responses across domain-specific agents coordinated by a central dialogue manager. It combines fine-tuned models (Flan-T5-large, Mistral-7B) with large LLMs (Claude Sonnet 3.0), yielding SOTA results on MultiWOZ with significant gains in inform and success rates. However, its performance depends heavily on extensive carefully designed prompt tuning and few-shot examples, limiting efficiency and increasing human effort.  \end{itemize}

\subsection{Human Study Details}

\begin{figure}[t]
  \centering
  \includegraphics[width=\linewidth]{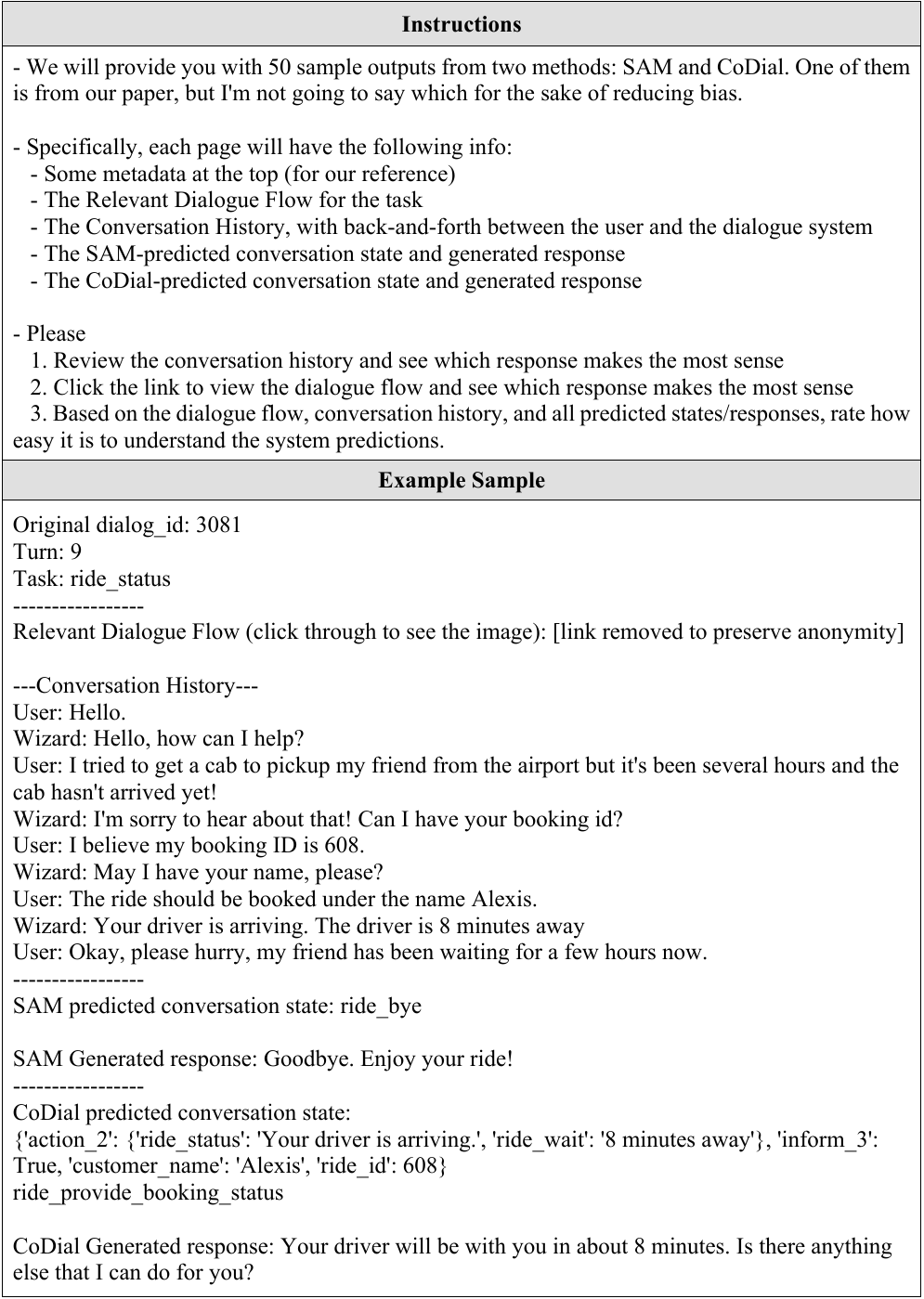}

   \caption{Full instructions from the human study, along with an example of the information provided to the annotator for one sample.}
   \label{fig:human-study-instructions}
\end{figure}

\label{ap:human-study}
To provide further evidence, we additionally conducted a human study with three human subjects. The three subject participants are student non-authors recruited by an internal call for participation, and have no prior knowledge of the research. We explicitly mentioned in the call that their responses would be used to assist in a publication. They are all graduate-level students with at least a Bachelor's degree, and have general knowledge of computer science/engineering with no prior exposure to both Colang and CoDial. We paid them \$20 per hour, slightly above the minimum wage in our jurisdiction.

We compare our CoDial method to one of our baselines, SAM. This was chosen over SGP-TOD because of the issues in reproducing SGP-TOD’s results (as discussed in \cref{ap:baselines}).

The three human subjects were shown conversation history, the SAM conversation state and output, and the CoDial conversation state and output, based on 50 randomly-selected conversation samples. They are then prompted with the following questions:

\begin{figure}[t]
    \centering
    \resizebox{0.99\linewidth}{!}{
    \includegraphics[width=\linewidth]{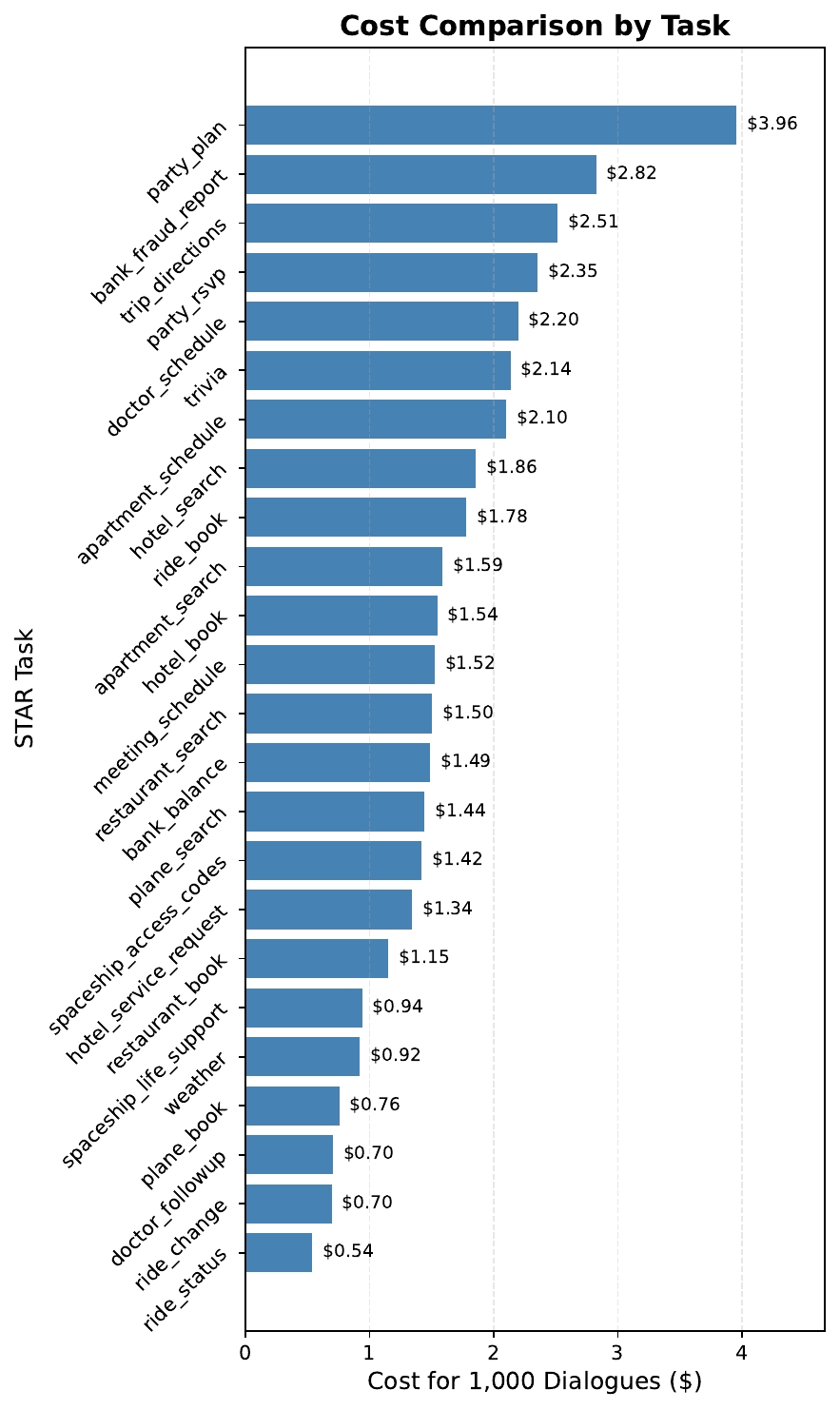}
    }
    \caption{Average CoDial (\texttt{4o, Qwen}) cost per 1{,}000 dialogues for 24 STAR tasks.}
    \label{fig:cost}
\end{figure}

\begin{itemize}
\item 
Q1. Which response makes more sense to the \textbf{conversation history}? (SAM / CoDial / Tie)
\item 
Q2. Which response makes more sense to the \textbf{dialogue flow}? (SAM / CoDial / Tie) 
\item 
Q3. How easy is it to understand the \textbf{SAM} output, including state and response? (Use a score between 1 to 5, with 1 meaning “Impossible to understand how the system arrived at the state shown” and 5 meaning “Most easy to understand how the system arrived at the state shown.”)
\item 
Q4. How easy is it to understand the \textbf{CoDial} output, including state and response? (Use a score between 1 to 5, with 1 meaning “Impossible to understand how the system arrived at the state shown” and 5 meaning “Most easy to understand how the system arrived at the state shown.”)
\end{itemize}

The annotators were not given any additional instructions, as we wanted to capture their subjective intuitions on what ``made sense'' and what was ``easy to understand.'' Questions 1 and 2 collect human preferences for the three choices. “Tie” means “no-preference”. The responses of the three human subjects are averaged to obtain the results. Questions 3 and 4 use a 5-point Likert scale, as detailed in the above questions. The results can be found in \Cref{tab:human-interpretability}.

\begin{figure*}[ht]
  \centering
  \includegraphics[width=\linewidth]{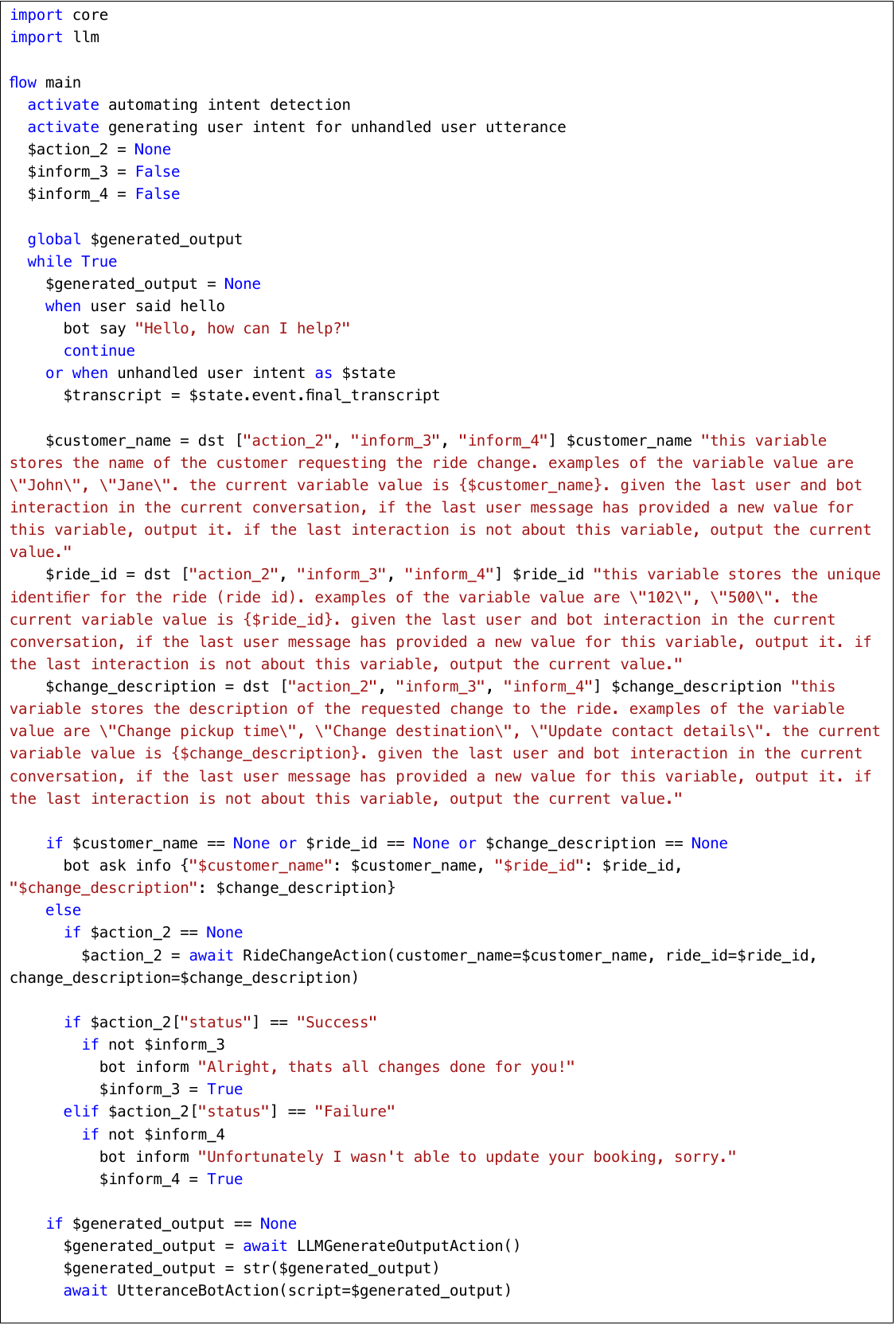}

   \caption{Example of a generated code for STAR \textit{Ride Change} task with \codials}
   \label{fig:example-code}
\end{figure*}

Additionally, the participants were shown a sample of CoDial code for the STAR's \textit{Ride Status} task. They were asked ``When CoDial's response doesn't make sense, how confident are you that you can fix the system's response? (1=not confident, 2=slightly confident, 3=moderately confident, 4=very confident, 5=absolutely confident).'' The average of the three subjects is 3.3, providing evidence showing that CoDial’s interpretable structure enables users to understand and improve the system’s behaviour---after encountering several faulty outputs, the human started to have confidence to use the intermediary guardrail structure to correct the underlying issues. An example of CoDial STAR's \textit{Ride Change} code can be found in \cref{fig:example-code}.

\begin{table}[t]
\centering
    \setlength{\tabcolsep}{8pt}
    \renewcommand\arraystretch{1.05}
    \resizebox{0.99\linewidth}{!}{
\begin{tabular}{lccc}
\toprule
\textbf{Metric} & \textbf{Overall} & \textbf{Min (per-task)} & \textbf{Max (per-task)} \\
\midrule
Tokens per dialogue & 19{,}291.0 & 6{,}370.0 & 47{,}565.0 \\
\quad input / output & 18{,}963.4 / 328.0 & 6{,}261.5 / 108.8 & 46{,}957.8 / 607.6 \\
Tokens per turn & 3{,}213.0 & 1{,}905.0 & 4{,}747.0 \\
\quad input / output & 3{,}158.8 / 54.6 & 1{,}872.6 / 32.5 & 4{,}686.8 / 60.6 \\
\midrule
Cost per 1{,}000 dialogues & \$1.63 & \$0.54 & \$3.96 \\
Cost per 1{,}000 turns     & \$0.27 & \$0.16 & \$0.38 \\
\bottomrule
\end{tabular}}
\caption{Cost analysis of CoDial on STAR for Qwen-3-30B-A3B-instruct. Token counts and costs are averaged per task; minimum and maximum are reported over 24 tasks.}
\label{tab:cost}
\end{table}

\begin{table}[t]
    \centering
    \setlength{\tabcolsep}{8pt}
    \renewcommand\arraystretch{1.05}
    \resizebox{0.8\linewidth}{!}{
    \begin{tabular}{lcc}
    \toprule
    \textbf{Method} & \textbf{Single-domain} & \textbf{Multi-domain} \\
    \midrule
    MARS    & 52.4 & 35.9 \scriptsize{(-16.5)} \\
    SOLOIST & 49.8 & 35.5 \scriptsize{(-14.3)} \\
    CoDial  & 46.2 & 28.4 \scriptsize{(-17.8)} \\
    \bottomrule
    \end{tabular}}

    \caption{Comparison of JGA for single- and multi-domain settings across different methods.} 
    \label{tab:mwz-single-vs-multi}
\end{table}

\begin{figure*}[ht]
  \centering
  \includegraphics[height=.9\textheight]{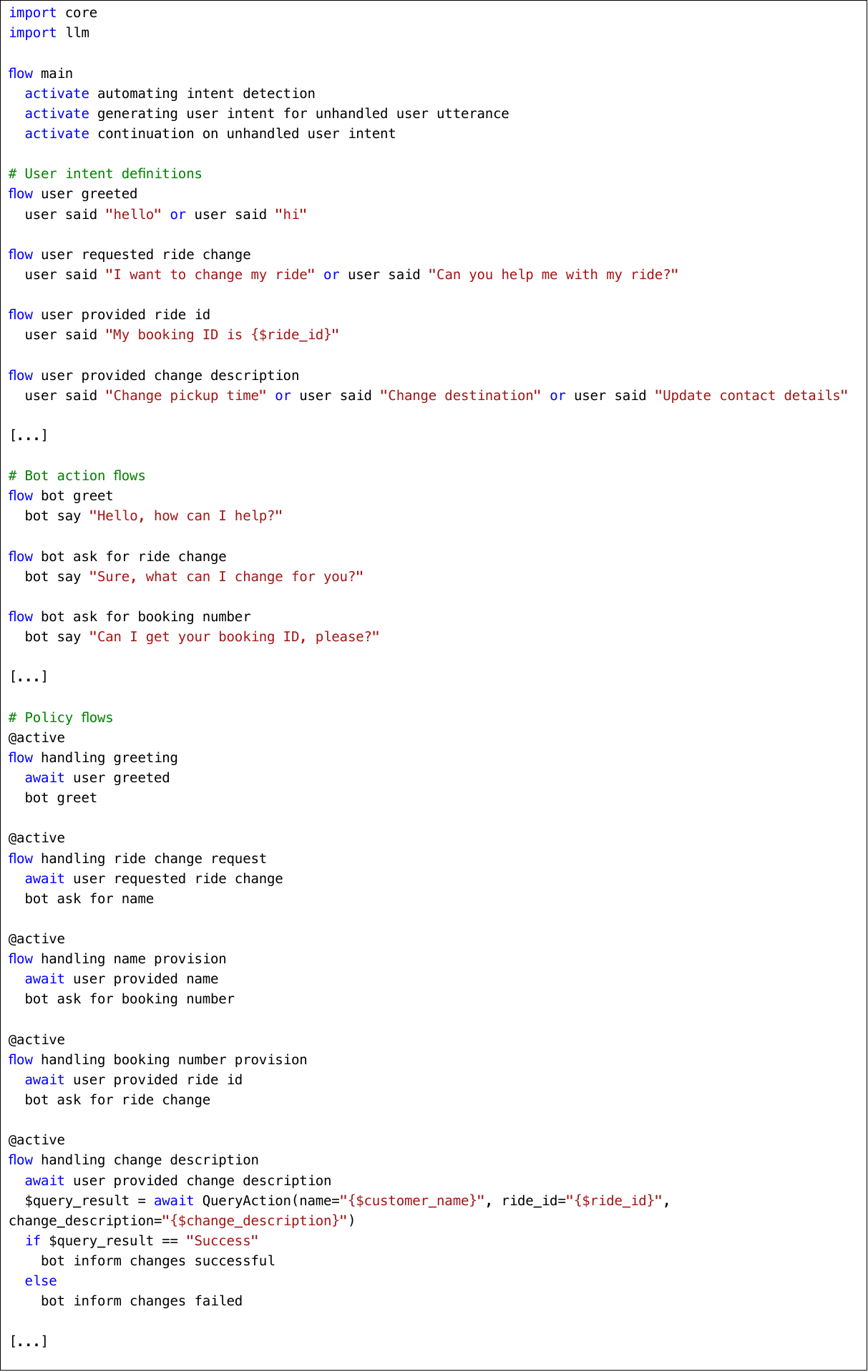}

   \caption{Example of a generated code for STAR \textit{Ride Change} task with \codialf}
   \label{fig:example-code-2}
\end{figure*}

%% file: custom.bib
@InProceedings{AnyTODProgrammableTask-zhao-2023,
  author    = {Zhao, Jeffrey and Cao, Yuan and Gupta, Raghav and Lee, Harrison and Rastogi, Abhinav and Wang, Mingqiu and Soltau, Hagen and Shafran, Izhak and Wu, Yonghui},
  booktitle = {Proceedings of the 2023 Conference on Empirical Methods in Natural Language Processing},
  title     = {{A}ny{TOD}: A Programmable Task-Oriented Dialog System},
  year      = {2023},
  address   = {Singapore},
  editor    = {Bouamor, Houda and Pino, Juan and Bali, Kalika},
  month     = dec,
  pages     = {16189--16204},
  publisher = {Association for Computational Linguistics},
  doi       = {10.18653/v1/2023.emnlp-main.1006},
  url       = {https://aclanthology.org/2023.emnlp-main.1006},
}

@InProceedings{SGPTODBuilding-zhang-2023,
  author    = {Zhang, Xiaoying and Peng, Baolin and Li, Kun and Zhou, Jingyan and Meng, Helen},
  booktitle = {Findings of the Association for Computational Linguistics: EMNLP 2023},
  title     = {{SGP}-{TOD}: Building Task Bots Effortlessly via Schema-Guided {LLM} Prompting},
  year      = {2023},
  address   = {Singapore},
  editor    = {Bouamor, Houda and Pino, Juan and Bali, Kalika},
  month     = dec,
  pages     = {13348--13369},
  publisher = {Association for Computational Linguistics},
  doi       = {10.18653/v1/2023.findings-emnlp.891},
  url       = {https://aclanthology.org/2023.findings-emnlp.891},
}

@Article{STARSchemaGuided-mosig-2020,
  author  = {Mosig, Johannes EM and Mehri, Shikib and Kober, Thomas},
  journal = {arXiv preprint arXiv:2010.11853},
  title   = {STAR: A Schema-Guided Dialog Dataset for Transfer Learning},
  year    = {2020},
}

@InProceedings{AreLargeLanguage-hudecek-2023,
  author    = {Hude{\v{c}}ek, Vojt{\v{e}}ch and Dusek, Ondrej},
  booktitle = {Proceedings of the 24th Annual Meeting of the Special Interest Group on Discourse and Dialogue},
  title     = {Are Large Language Models All You Need for Task-Oriented Dialogue?},
  year      = {2023},
  address   = {Prague, Czechia},
  editor    = {Stoyanchev, Svetlana and Joty, Shafiq and Schlangen, David and Dusek, Ondrej and Kennington, Casey and Alikhani, Malihe},
  month     = sep,
  pages     = {216--228},
  publisher = {Association for Computational Linguistics},
  doi       = {10.18653/v1/2023.sigdial-1.21},
  url       = {https://aclanthology.org/2023.sigdial-1.21},
}

@InProceedings{Doyoufollow-jacqmin-2022,
  author    = {Jacqmin, L{\'e}o and Rojas Barahona, Lina M. and Favre, Benoit},
  booktitle = {Proceedings of the 23rd Annual Meeting of the Special Interest Group on Discourse and Dialogue},
  title     = {{``}Do you follow me?{''}: A Survey of Recent Approaches in Dialogue State Tracking},
  year      = {2022},
  address   = {Edinburgh, UK},
  editor    = {Lemon, Oliver and Hakkani-Tur, Dilek and Li, Junyi Jessy and Ashrafzadeh, Arash and Garcia, Daniel Hern{\'a}ndez and Alikhani, Malihe and Vandyke, David and Du{\v{s}}ek, Ond{\v{r}}ej},
  month     = sep,
  pages     = {336--350},
  publisher = {Association for Computational Linguistics},
  doi       = {10.18653/v1/2022.sigdial-1.33},
  url       = {https://aclanthology.org/2022.sigdial-1.33},
}

@InProceedings{SchemaGuidedParadigm-mehri-2021,
  author    = {Mehri, Shikib and Eskenazi, Maxine},
  booktitle = {Proceedings of the 22nd Annual Meeting of the Special Interest Group on Discourse and Dialogue},
  title     = {Schema-Guided Paradigm for Zero-Shot Dialog},
  year      = {2021},
  address   = {Singapore and Online},
  editor    = {Li, Haizhou and Levow, Gina-Anne and Yu, Zhou and Gupta, Chitralekha and Sisman, Berrak and Cai, Siqi and Vandyke, David and Dethlefs, Nina and Wu, Yan and Li, Junyi Jessy},
  month     = jul,
  pages     = {499--508},
  publisher = {Association for Computational Linguistics},
  doi       = {10.18653/v1/2021.sigdial-1.52},
  url       = {https://aclanthology.org/2021.sigdial-1.52},
}

@InProceedings{NeMoGuardrailsToolkit-rebedea-2023,
  author    = {Rebedea, Traian and Dinu, Razvan and Sreedhar, Makesh Narsimhan and Parisien, Christopher and Cohen, Jonathan},
  booktitle = {Proceedings of the 2023 Conference on Empirical Methods in Natural Language Processing: System Demonstrations},
  title     = {{N}e{M}o Guardrails: A Toolkit for Controllable and Safe {LLM} Applications with Programmable Rails},
  year      = {2023},
  address   = {Singapore},
  editor    = {Feng, Yansong and Lefever, Els},
  month     = dec,
  pages     = {431--445},
  publisher = {Association for Computational Linguistics},
  doi       = {10.18653/v1/2023.emnlp-demo.40},
  url       = {https://aclanthology.org/2023.emnlp-demo.40},
}

@Misc{BuildingGuardrailsLarge-dong-2024,
  author        = {Yi Dong and Ronghui Mu and Gaojie Jin and Yi Qi and Jinwei Hu and Xingyu Zhao and Jie Meng and Wenjie Ruan and Xiaowei Huang},
  title         = {Building Guardrails for Large Language Models},
  year          = {2024},
  archiveprefix = {arXiv},
  eprint        = {2402.01822},
  primaryclass  = {cs.CL},
  url           = {https://arxiv.org/abs/2402.01822},
}

@article{chen2017survey,
  title={A survey on dialogue systems: Recent advances and new frontiers},
  author={Chen, Hongshen and Liu, Xiaorui and Yin, Dawei and Tang, Jiliang},
  journal={Acm Sigkdd Explorations Newsletter},
  volume={19},
  number={2},
  pages={25--35},
  year={2017},
  publisher={Acm New York, NY, USA}
}

@article{liu2024your,
  title={Is your code generated by chatgpt really correct? rigorous evaluation of large language models for code generation},
  author={Liu, Jiawei and Xia, Chunqiu Steven and Wang, Yuyao and Zhang, Lingming},
  journal={Advances in Neural Information Processing Systems},
  volume={36},
  year={2024}
}

@inproceedings{NEURIPS2022_8636419d,
 author = {Le, Hung and Wang, Yue and Gotmare, Akhilesh Deepak and Savarese, Silvio and Hoi, Steven Chu Hong},
 booktitle = {Advances in Neural Information Processing Systems},
 editor = {S. Koyejo and S. Mohamed and A. Agarwal and D. Belgrave and K. Cho and A. Oh},
 pages = {21314--21328},
 publisher = {Curran Associates, Inc.},
 title = {CodeRL: Mastering Code Generation through Pretrained Models and Deep Reinforcement Learning},
 url = {https://proceedings.neurips.cc/paper_files/paper/2022/file/8636419dea1aa9fbd25fc4248e702da4-Paper-Conference.pdf},
 volume = {35},
 year = {2022}
}

@article{jiang2024self,
  title={Self-planning code generation with large language models},
  author={Jiang, Xue and Dong, Yihong and Wang, Lecheng and Fang, Zheng and Shang, Qiwei and Li, Ge and Jin, Zhi and Jiao, Wenpin},
  journal={ACM Transactions on Software Engineering and Methodology},
  volume={33},
  number={7},
  pages={1--30},
  year={2024},
  publisher={ACM New York, NY}
}

@InProceedings{TowardsLLMdriven-feng-2023,
  author    = {Feng, Yujie and Lu, Zexin and Liu, Bo and Zhan, Liming and Wu, Xiao-Ming},
  booktitle = {Proceedings of the 2023 Conference on Empirical Methods in Natural Language Processing},
  title     = {Towards {LLM}-driven Dialogue State Tracking},
  year      = {2023},
  address   = {Singapore},
  editor    = {Bouamor, Houda and Pino, Juan and Bali, Kalika},
  month     = dec,
  pages     = {739--755},
  publisher = {Association for Computational Linguistics},
  doi       = {10.18653/v1/2023.emnlp-main.48},
  url       = {https://aclanthology.org/2023.emnlp-main.48/},
}

@InProceedings{ShadesBLEUFlavours-nekvinda-2021,
  author    = {Nekvinda, Tom{\'a}{\v{s}} and Du{\v{s}}ek, Ond{\v{r}}ej},
  booktitle = {Proceedings of the 1st Workshop on Natural Language Generation, Evaluation, and Metrics (GEM 2021)},
  title     = {Shades of {BLEU}, Flavours of Success: The Case of {M}ulti{WOZ}},
  year      = {2021},
  address   = {Online},
  editor    = {Bosselut, Antoine and Durmus, Esin and Gangal, Varun Prashant and Gehrmann, Sebastian and Jernite, Yacine and Perez-Beltrachini, Laura and Shaikh, Samira and Xu, Wei},
  month     = aug,
  pages     = {34--46},
  publisher = {Association for Computational Linguistics},
  doi       = {10.18653/v1/2021.gem-1.4},
  url       = {https://aclanthology.org/2021.gem-1.4/},
}

@inproceedings{qin-etal-2023-end,
    title = "End-to-end Task-oriented Dialogue: A Survey of Tasks, Methods, and Future Directions",
    author = "Qin, Libo  and
      Pan, Wenbo  and
      Chen, Qiguang  and
      Liao, Lizi  and
      Yu, Zhou  and
      Zhang, Yue  and
      Che, Wanxiang  and
      Li, Min",
    editor = "Bouamor, Houda  and
      Pino, Juan  and
      Bali, Kalika",
    booktitle = "Proceedings of the 2023 Conference on Empirical Methods in Natural Language Processing",
    month = dec,
    year = "2023",
    address = "Singapore",
    publisher = "Association for Computational Linguistics",
    url = "https://aclanthology.org/2023.emnlp-main.363/",
    doi = "10.18653/v1/2023.emnlp-main.363",
    pages = "5925--5941"
}

@InProceedings{MultiWOZLargeScale-budzianowski-2018,
  author    = {Budzianowski, Pawe{\l}  and
      Wen, Tsung-Hsien  and
      Tseng, Bo-Hsiang  and
      Casanueva, I{\~n}igo  and
      Ultes, Stefan  and
      Ramadan, Osman  and
      Ga{\v{s}}i{\'c}, Milica},
  booktitle = {Proceedings of the 2018 Conference on Empirical Methods in Natural Language Processing},
  title     = {{M}ulti{WOZ} - A Large-Scale Multi-Domain {W}izard-of-{O}z Dataset for Task-Oriented Dialogue Modelling},
  year      = {2018},
  address   = {Brussels, Belgium},
  editor    = {Riloff, Ellen  and
      Chiang, David  and
      Hockenmaier, Julia  and
      Tsujii, Jun{'}ichi},
  month     = oct # {-} # nov,
  pages     = {5016--5026},
  publisher = {Association for Computational Linguistics},
  doi       = {10.18653/v1/D18-1547},
  url       = {https://aclanthology.org/D18-1547/},
}

@InProceedings{BLEUmethodautomatic-papineni-2002,
  author    = {Papineni, Kishore and Roukos, Salim and Ward, Todd and Zhu, Wei-Jing},
  booktitle = {Proceedings of the 40th Annual Meeting on Association for Computational Linguistics},
  title     = {BLEU: a method for automatic evaluation of machine translation},
  year      = {2002},
  address   = {USA},
  pages     = {311–318},
  publisher = {Association for Computational Linguistics},
  series    = {ACL '02},
  doi       = {10.3115/1073083.1073135},
  location  = {Philadelphia, Pennsylvania},
  numpages  = {8},
  url       = {https://doi.org/10.3115/1073083.1073135},
}

@InProceedings{CallClarityReporting-post-2018,
  author    = {Post, Matt},
  booktitle = {Proceedings of the Third Conference on Machine Translation: Research Papers},
  title     = {A Call for Clarity in Reporting {BLEU} Scores},
  year      = {2018},
  address   = {Belgium, Brussels},
  month     = oct,
  pages     = {186--191},
  publisher = {Association for Computational Linguistics},
  url       = {https://www.aclweb.org/anthology/W18-6319},
}

@InProceedings{FnCTOD-li-2024,
  author    = {Li, Zekun  and
      Chen, Zhiyu  and
      Ross, Mike  and
      Huber, Patrick  and
      Moon, Seungwhan  and
      Lin, Zhaojiang  and
      Dong, Xin  and
      Sagar, Adithya  and
      Yan, Xifeng  and
      Crook, Paul},
  booktitle = {Proceedings of the 62nd Annual Meeting of the Association for Computational Linguistics (Volume 1: Long Papers)},
  title     = {Large Language Models as Zero-shot Dialogue State Tracker through Function Calling},
  year      = {2024},
  address   = {Bangkok, Thailand},
  editor    = {Ku, Lun-Wei  and
      Martins, Andre  and
      Srikumar, Vivek},
  month     = aug,
  pages     = {8688--8704},
  publisher = {Association for Computational Linguistics},
  doi       = {10.18653/v1/2024.acl-long.471},
  url       = {https://aclanthology.org/2024.acl-long.471/},
}

@article{yang2023large,
  title={Large language models as optimizers},
  author={Yang, Chengrun and Wang, Xuezhi and Lu, Yifeng and Liu, Hanxiao and Le, Quoc V and Zhou, Denny and Chen, Xinyun},
  journal={arXiv preprint arXiv:2309.03409},
  year={2023}
}

@inproceedings{ji2023towards,
  title={Towards mitigating LLM hallucination via self reflection},
  author={Ji, Ziwei and Yu, Tiezheng and Xu, Yan and Lee, Nayeon and Ishii, Etsuko and Fung, Pascale},
  booktitle={Findings of the Association for Computational Linguistics: EMNLP 2023},
  pages={1827--1843},
  year={2023}
}

@article{yuksekgonul2024textgrad,
      title={TextGrad: Automatic "Differentiation" via Text},
      author={Mert Yuksekgonul and Federico Bianchi and Joseph Boen and Sheng Liu and Zhi Huang and Carlos Guestrin and James Zou},
      year={2024},
      eprint={2406.07496},
      archivePrefix={arXiv}
}

@article{dahan2023lawyers,
  title={Lawyers Should Not Trust AI: A call for an Open-source Legal Language Model},
  author={Dahan, Samuel and Bhambhoria, Rohan and Liang, David and Zhu, Xiaodan},
  journal={Available at SSRN 4587092},
  year={2023}
}

@article{tian2024opportunities,
  title={Opportunities and challenges for ChatGPT and large language models in biomedicine and health},
  author={Tian, Shubo and Jin, Qiao and Yeganova, Lana and Lai, Po-Ting and Zhu, Qingqing and Chen, Xiuying and Yang, Yifan and Chen, Qingyu and Kim, Won and Comeau, Donald C and others},
  journal={Briefings in Bioinformatics},
  volume={25},
  number={1},
  pages={bbad493},
  year={2024},
  publisher={Oxford University Press}
}

@Article{SoloistBuildingTask-peng-2021,
  author    = {Peng, Baolin  and
      Li, Chunyuan  and
      Li, Jinchao  and
      Shayandeh, Shahin  and
      Liden, Lars  and
      Gao, Jianfeng},
  journal   = {Transactions of the Association for Computational Linguistics},
  title     = {Soloist: Building Task Bots at Scale with Transfer Learning and Machine Teaching},
  year      = {2021},
  pages     = {807--824},
  volume    = {9},
  address   = {Cambridge, MA},
  doi       = {10.1162/tacl_a_00399},
  editor    = {Roark, Brian  and
      Nenkova, Ani},
  publisher = {MIT Press},
  url       = {https://aclanthology.org/2021.tacl-1.49/},
}

@InProceedings{MarsModelingContext-sun-2023,
  author    = {Sun, Haipeng  and
      Bao, Junwei  and
      Wu, Youzheng  and
      He, Xiaodong},
  booktitle = {Findings of the Association for Computational Linguistics: ACL 2023},
  title     = {{M}ars: Modeling Context {\&} State Representations with Contrastive Learning for End-to-End Task-Oriented Dialog},
  year      = {2023},
  address   = {Toronto, Canada},
  editor    = {Rogers, Anna  and
      Boyd-Graber, Jordan  and
      Okazaki, Naoaki},
  month     = jul,
  pages     = {11139--11160},
  publisher = {Association for Computational Linguistics},
  doi       = {10.18653/v1/2023.findings-acl.708},
  url       = {https://aclanthology.org/2023.findings-acl.708/},
}

@Misc{DeepSeekV3Technical-deepseek-ai-2024,
  author        = {DeepSeek-AI and Aixin Liu and Bei Feng and Bing Xue and Bingxuan Wang and Bochao Wu and Chengda Lu and Chenggang Zhao and Chengqi Deng and Chenyu Zhang and Chong Ruan and others},
  title         = {DeepSeek-V3 Technical Report},
  year          = {2024},
  archiveprefix = {arXiv},
  eprint        = {2412.19437},
  primaryclass  = {cs.CL},
  url           = {https://arxiv.org/abs/2412.19437},
}

@misc{guardrailsai,
  title = {Guardrails: Adding Guardrails to Large Language Models},
  author = {{Guardrails AI}},
  howpublished = {\url{https://github.com/guardrails-ai/guardrails}},
  note = {Accessed: 2025-05-16}
}

@online{nvidia2024nemo,
  author       = {NVIDIA},
  title        = {NVIDIA NeMo Guardrails, docs.nvidia.com/nemo/guardrails/colang\_2/overview .html},
  year         = {2024},
  url          = {https://docs.nvidia.com/nemo/guardrails/colang_2/overview.html},
  note         = {Accessed: 2025-05-19}
}

@InProceedings{DARDMultiAgent-gupta-2024,
  author    = {Aman Gupta and Anirudh Ravichandran and Narayanan Sadagopan and Anurag Beniwal},
  booktitle = {NeurIPS 2024 Workshop on Open-World Agents},
  title     = {{DARD}: A Multi-Agent Approach for Task-Oriented Dialog Systems},
  year      = {2024},
  url       = {https://openreview.net/forum?id=RbkX9e4qqP},
}

@Article{WorkflowDiscoveryDialogues-hattami-2023,
  author  = {Amine El Hattami and Issam H. Laradji and Stefania Raimondo and David Vazquez and Pau Rodriguez and Christopher Pal},
  journal = {Transactions on Machine Learning Research},
  title   = {Workflow Discovery from Dialogues in the Low Data Regime},
  year    = {2023},
  issn    = {2835-8856},
  note    = {Featured Certification},
  url     = {https://openreview.net/forum?id=L9othQvPks},
}

@InProceedings{UnsupervisedExtractionDialogue-sreedhar-2024,
  author    = {Sreedhar, Makesh Narsimhan and Rebedea, Traian and Parisien, Christopher},
  booktitle = {Proceedings of the 2024 Conference on Empirical Methods in Natural Language Processing},
  title     = {Unsupervised Extraction of Dialogue Policies from Conversations},
  year      = {2024},
  address   = {Miami, Florida, USA},
  editor    = {Al-Onaizan, Yaser and Bansal, Mohit and Chen, Yun-Nung},
  month     = nov,
  pages     = {19029--19045},
  publisher = {Association for Computational Linguistics},
  doi       = {10.18653/v1/2024.emnlp-main.1060},
  url       = {https://aclanthology.org/2024.emnlp-main.1060/},
}

@InProceedings{UnsupervisedLearningHierarchical-lu-2022,
  author    = {Lu, Bo-Ru and Hu, Yushi and Cheng, Hao and Smith, Noah A. and Ostendorf, Mari},
  booktitle = {Findings of the Association for Computational Linguistics: EMNLP 2022},
  title     = {Unsupervised Learning of Hierarchical Conversation Structure},
  year      = {2022},
  address   = {Abu Dhabi, United Arab Emirates},
  editor    = {Goldberg, Yoav and Kozareva, Zornitsa and Zhang, Yue},
  month     = dec,
  pages     = {5657--5670},
  publisher = {Association for Computational Linguistics},
  doi       = {10.18653/v1/2022.findings-emnlp.415},
  url       = {https://aclanthology.org/2022.findings-emnlp.415/},
}

@article{dong2024safeguarding,
  title={Safeguarding large language models: a survey. arXiv},
  author={Dong, Y and Mu, R and Zhang, Y and Sun, S and Zhang, T and Wu, C and Jin, G and Qi, Y and Hu, J and Meng, J},
  journal={Preprint posted online June},
  volume={3},
  year={2024}
}

@article{yang2025qwen3,
  title={Qwen3 technical report},
  author={Yang, An and Li, Anfeng and Yang, Baosong and Zhang, Beichen and Hui, Binyuan and Zheng, Bo and Yu, Bowen and Gao, Chang and Huang, Chengen and Lv, Chenxu and others},
  journal={arXiv preprint arXiv:2505.09388},
  year={2025}
}
